\documentclass[notitlepage]{article}
\usepackage[T1]{fontenc}
\usepackage[utf8]{inputenc}

\usepackage{booktabs}
\usepackage[authoryear]{natbib}
\usepackage{graphicx}
\usepackage{authblk}

\date{}

\usepackage[font=small,labelfont=bf,labelsep=period]{caption}
\usepackage{subcaption}
\usepackage[bottom]{footmisc}
\usepackage{wrapfig}
\usepackage{float}
\usepackage{hyperref}
\usepackage{enumitem}
\usepackage{amsmath}
\usepackage{amssymb}
\usepackage{sansmath}
\usepackage{bm}

\newcommand{\weights}{w}
\numberwithin{equation}{section}

\usepackage{dsfont}

\usepackage{multirow}
\usepackage{cellspace}
\usepackage{hhline}
\usepackage{stackengine}
\setstackEOL{\\}
\usepackage[misc]{ifsym}
\usepackage{amsthm}
\usepackage{amsmath}
\let\svparbox\parbox
\renewcommand\parbox[3][c]{\svparbox[#1]{#2}{\strut#3\strut}}
\usepackage{caption}
\newtheorem*{assump*}{Assumption}
\theoremstyle{remark}

\usepackage{xcolor}

\usepackage{mathpazo}
\usepackage[scaled=0.90]{helvet}
\usepackage[scaled=0.85]{beramono}

\makeatletter
\newcommand{\printfnsymbol}[1]{%
  \textsuperscript{\@fnsymbol{#1}}%
}
\makeatother

\title{Neural Topic Models with Survival Supervision: Jointly Predicting Time-to-Event Outcomes and Learning How Clinical Features Relate}

\author[1]{George H.~Chen\thanks{equal contribution}}
\author[2]{Linhong Li\printfnsymbol{1}}
\author[3]{Ren Zuo}
\author[4]{Amanda Coston}
\author[5]{Jeremy C.~Weiss}
\affil[1]{Heinz College of Information Systems and Public Policy, Carnegie Mellon University}
\affil[2]{McKinsey \& Company}
\affil[3]{Cornerstone Research}
\affil[4]{Microsoft Research}
\affil[5]{National Library of Medicine, National Institutes of Health}

\begin{document}

\sloppy

\maketitle{}
\vspace{-2em}

\begin{abstract}
We present a neural network framework for learning a survival model to predict a time-to-event outcome while simultaneously learning a topic model that reveals feature relationships. In particular, we model each subject as a distribution over ``topics'', where a topic could, for instance, correspond to an age group, a disorder, or a disease. The presence of a topic in a subject means that specific clinical features are more likely to appear for the subject. Topics encode information about related features and are learned in a supervised manner to predict a time-to-event outcome. Our framework supports combining many different topic and survival models; training the resulting joint survival-topic model readily scales to large datasets using standard neural net optimizers with minibatch gradient descent. For example, a special case is to combine LDA with a Cox model, in which case a subject's distribution over topics serves as the input feature vector to the Cox model. We explain how to address practical implementation issues that arise when applying these neural survival-supervised topic models to clinical data, including how to visualize results to assist clinical interpretation. We study the effectiveness of our proposed framework on seven clinical datasets on predicting time until death as well as hospital ICU length of stay, where we find that neural survival-supervised topic models achieve competitive accuracy with existing approaches while yielding interpretable clinical topics that explain feature relationships. Our code is available at: \url{https://github.com/georgehc/survival-topics}
\end{abstract}

\section{Introduction}
\label{sec:intro}

Predicting the amount of time until a critical event occurs---such as death, disease relapse, or hospital discharge---is a central focus in the field of survival analysis. Especially with the increasing availability of electronic health records, survival analysis data in healthcare often have both a large number of subjects and a large number of features measured per subject. In coming up with an interpretable survival analysis model to predict time-to-event outcomes for these large-scale datasets, a standard approach is to use the classical Cox proportional hazards model \citep{cox1972regression}, possibly with features selected using lasso regularization \citep{simon2011regularization} or stepwise regression \citep{harrell1984regression}. However, these Cox-based models do not inherently learn how features relate. Instead, to try to understand feature interactions with a Cox model, one would have to, for example, introduce a large number of features that encode interactions between the original features. This approach is impractical when the number of features is very large.

To simultaneously address the two objectives of learning a survival model for time-to-event prediction and learning how features relate through a topic model, \citet{dawson2012survival} combine latent Dirichlet allocation (LDA) \citep{blei2003latent} with Cox proportional hazards to obtain a method they call \textsc{survLDA}. The idea is to represent each subject as a distribution over topics, and each topic as a distribution over which clinical feature values appear. For example, a topic could correspond to a severe disease state or a particular age group. The Cox model is given the subjects' distributions over topics as input rather than the subjects' raw feature vectors. Importantly, the topic and survival models are jointly learned.

In this paper, we propose a general framework for deriving neural survival-supervised topic models that is substantially more flexible than \textsc{survLDA}. Specifically, \textsc{survLDA} estimates model parameters via variational inference update equations derived specifically for LDA combined with the standard Cox model; to use another other sort of combination would require re-deriving the inference algorithm. Moreover, the inference algorithm for \textsc{survLDA} as stated in their paper does not easily scale to large datasets. In contrast, our approach combines essentially any topic model and any survival model that can be cast in a neural net framework (precise prerequisites of our framework are given in Section~\ref{sec:background}); combining LDA with the Cox proportional hazards model is only one special case. As a byproduct of taking a neural net approach, we can readily leverage many deep learning advances. For example, we can avoid deriving a special inference algorithm and instead use any neural net optimizer such as Adam \citep{kingma2015adam} to learn the joint model in minibatches, which readily scales to large datasets. Importantly, our framework yields survival-supervised topic models that are amenable to interpretation so long as the underlying topic and survival models are.

As numerous combinations of neural topic/survival models are possible, we only demonstrate four combinations, corresponding to combining either LDA or SAGE \citep{eisenstein2011sparse} topic models with either the Cox proportional hazards model or an accelerated failure time model (e.g., \citealt{cox1972regression,prentice1978linear}). We make these combinations within the \textsc{scholar} neural topic modeling framework by \citet{card2018neural} and thus refer to the resulting neural survival-supervised topic models as \textsc{scholar lda-cox}, \textsc{scholar lda-aft}, \textsc{scholar sage-cox}, and \textsc{scholar sage-aft}; note that \textsc{scholar lda-cox} is a neural network variant of \textsc{survLDA}. We benchmark the four neural survival-supervised models on seven datasets, finding that they can yield accuracy competitive with deep learning baselines \citep{katzman2018deepsurv,lee2018deephit} while yielding interpretable topics. In contrast, the deep learning baselines are not interpretable.

Importantly, we discuss practical challenges encountered in learning these neural survival-supervised topic models on clinical data to obtain interpretable topics. For example, we found the standard approach in topic modeling of just listing the top features per topic to often not be interpretable because this listing does not explain how these top features' probabilities of appearing vary across topics. As an alternative, we propose a new heatmap visualization of learned topics that we found can better assist clinical interpretation. Separately, we find encouraging sparsity in learned topics to make the topics \emph{less} interpretable. Our observation is that sometimes multiple clinical events/measurements are taken that altogether help explain a condition, whereas encouraging sparsity tends to only pick out one among multiple related features. This is essentially the same problem encountered when using lasso for linear regression: when there is a group of variables with high pairwise correlation, lasso arbitrarily chooses one of these variables \citep{zou2005regularization}. We do not want this sort of behavior when our goal is to understand how different features relate.

As a separate issue on interpretability, especially when the number of features is large, it is possible that many features do not help explain survival outcomes. \citet{dawson2012survival} address this issue by using a preprocessing procedure for \textsc{survLDA}. Specifically, they cluster on the subjects' data based on their survival outcomes. Then they remove features that are not sufficiently different across the clusters. The issue with this approach is that it is ad hoc and how it impacts downstream analyses is unclear. Moreover, there are many possible clustering approaches that can be used each with its own (hyper)parameter settings that can be tuned. We do not use such a heuristic preprocessing step to filter features. Instead, we filter features \emph{after} learning a survival-supervised topic model. This strategy has been demonstrated to work as well as filtering features \emph{before} learning topic models \citep{schofield2017pulling} although it has not been demonstrated in the survival analysis context. Filtering after learning the model is appealing since we can apply different filters (potentially with clinician input) without having to retrain the model. For example, we can screen out features that appear in too few or too many patients on demand after learning the model.

\begin{figure}
\centering
\includegraphics[scale=0.67]{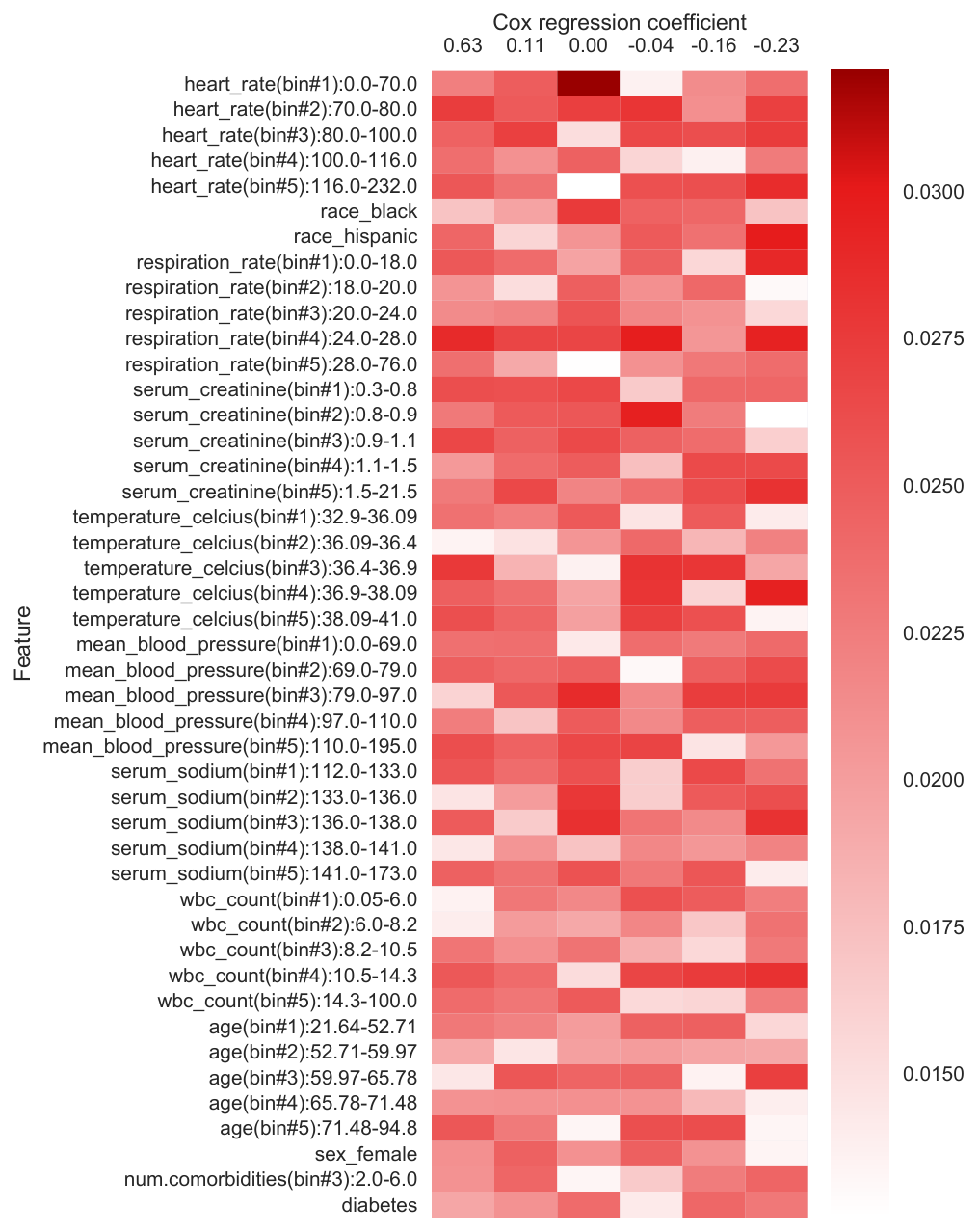}
\caption{Topics learned by \textsc{scholar lda-cox} on the \textsc{support3} (cancer) dataset. Columns index topics and rows index features/``words''. The values are probabilities of each feature conditioned on being in a topic. Note that two different features that are highly probable (darker shade of red) for the same topic does \emph{not} mean that they must co-occur when that topic is present, and it is possible that neither occurs. A helpful way to think about this is to consider how topic modeling works when applied to text data such as news articles. In this case, a learned topic might correspond to \emph{sports}, which could have highly probable words such as ``basketball'' and ``skiing''. A text document could be about sports yet mentions neither of these words. This same idea applies to our setting where we represent patients in terms of clinical topics.}
\label{fig:support3-heatmap}
\end{figure}

As a concrete example, on a cancer dataset where we aim to predict time until death, the topics learned by one of our neural survival-supervised topic models \textsc{scholar lda-cox} are shown as a heatmap in Figure~\ref{fig:support3-heatmap}. In the heatmap, the columns correspond to different topics (ordered from left to right corresponding to being associated with shorter to longer average survival time), the rows correspond to different clinical measurements (continuous measurements are discretized into bins), and the color values are probabilities where a deeper red roughly means that the feature is more prominent for a particular topic. We explain in Section~\ref{sec:experiments} precisely how this heatmap is constructed and how the rows are ordered. By looking at this heatmap, we can quickly identify how feature occurrences tend to differ across the topics. We can interpret the topics by looking at which features tend to be highly probable for each topic. Our resulting interpretations are shown in Table~\ref{tab:support3-scholar-lda-cox-topic-summary}.

Extremely importantly, the interpretation of the learned topics requires an abundance of caution. While our learned topic models are competitive with various state-of-the-art baselines in terms of prediction accuracy, the best accuracy scores possible are not high for the various prediction tasks we consider in our experiments. Thus, we cannot claim that the learned topics are ``correct'', and we believe that they require more extensive validation if they are to be deployed for clinical use. However, the learned topics can be very helpful in model debugging. By visualizing them with our heatmap strategy, we can spot inconsistencies between topics learned and clinical intuition, which could suggest ways to improve the model (e.g., adding constraints or regularization, changing specific data preprocessing steps). In contrast, state-of-the-art deep learning baselines that we benchmark against are not interpretable and do not provide straightforward visualizations to assist model debugging and improvement.

With the above disclaimer, if we suppose for the moment that the learned topics in Figure~\ref{fig:support3-heatmap}/Table~\ref{tab:support3-scholar-lda-cox-topic-summary} capture valid associations, then the topics could provide actionable insights. In the problem of predicting time until death for cancer patients, we may want to tease apart elderly cancer patients in terms of their risk of mortality. Topics 1, 4, and 5 (as numbered in Table~\ref{tab:support3-scholar-lda-cox-topic-summary}) would be particularly relevant in this case as they focus more on elderly patients and are associated with different risks of mortality. By looking at what differentiates these topics, we see that fever, infection, and inflammation are key indicators, which we could consider interventions for. Note that whether a patient is more associated with topic 1 vs 5 can be distinguished by other characteristics such as blood pressure and white blood cell count. One might want to consider more aggressive interventions for patients mostly associated with topic 1 since their prognosis is worse collectively.

\begin{table}
\caption{Summary of topics learned by \textsc{scholar lda-cox} on the \textsc{support3} (cancer) dataset. Higher Cox regression $\beta$ coefficient is associated with shorter survival time.}
\label{tab:support3-scholar-lda-cox-topic-summary}
\centering
\footnotesize
{\renewcommand{\arraystretch}{1.2}
\begin{tabular}{ccp{0.6\linewidth}}\toprule
Topic number & $\beta$ & Topic interpretation \\\midrule
1 & 0.63 & old otherwise normal \\
2 & 0.11 & cardiorenal problems with comorbidities \\
3 & 0 & baseline \\
4 & $-0.04$ & old, feverish, infection/inflammation \\
5 & $-0.16$ & old with inflammation \\
6 & $-0.23$ & normal healthier \\\bottomrule
\end{tabular}
}
\end{table}

In summary, our main contributions are as follows:
\begin{itemize}
\item We propose a general neural network framework for combining neural topic models with survival models. This framework is meant for large datasets in which both the number of subjects and the number of features are large, where a key goal is to discover possible feature relationships.
\item We discuss practical issues that arise when applying our framework to clinical data, including visualization strategies to assist clinical interpretation.
\item We experimentally show that neural survival-supervised topic models often work as well as deep learning baselines but have the added advantage of producing clinically interpretable topics. The deep learning baselines are not interpretable.
\end{itemize}
\paragraph{Outline} The rest of the paper is organized as follows. We provide background and prerequisites of our framework in Section~\ref{sec:background}. We then explain how to construct neural-survival supervised topic models with an explicit background topic in Section~\ref{sec:survival-scholar}, with examples given for how to combine LDA and SAGE topic models with the Cox and log-logistic accelerated failure time survival models. We then benchmark these models against classical and deep learning baselines in Section~\ref{sec:experiments}, where we also discuss model interpretability. We end the paper with a discussion in Section~\ref{sec:discussion}.

\section{Background and Prerequisites for Our Framework}
\label{sec:background}

We begin with some background and notation, first stating the format of the data we assume we have access to. Then we review key ideas of topic modeling and survival analysis most pertinent to our proposed framework. Importantly, we state what properties our framework requires of the topic and survival models that will be combined to form a neural survival-supervised topic model. For ease of exposition, we phrase notation in terms of predicting time until death; other critical events are possible aside from death.

\subsection{Data Format}

We assume that we have access to a training dataset of $n$ subjects, and we pre-specify $d$ historical clinical events to keep track of, where each event either occurs or not. For example, a clinical event could be whether a patient was ever diagnosed with diabetes up to present time. Continuous-valued clinical measurements could be discretized into bins to come up with such binary historical clinical events. For example, white blood count could be discretized into five quintiles. Thus, one of the $d$ events would then be ``white blood count reading is in the bottom quintile''; this event could occur multiple times. For a given subject, we can count how many times each of the $d$ events happened up to present time. We denote $X_{i,u}$ to be the number of times event $u\in\{1,\dots,d\}$ occurred for subject $i \in \{1,\dots,n\}$.\footnote{For simplicity, especially as the focus of our paper is not on feature engineering or preprocessing (which often needs to be tailored to specific datasets), when working with continuous-valued features, we use the simple quintile binning strategy we described along with counting how often each discretized event occurs across time to obtain the raw counts matrix $X$. In practice, one could of course use other discretization strategies, whether based on known threshold values that are already in clinical use for specific features, or based on automatically learned threshold values. Moreover, rather than counting how often a (discretized) measurement occurs over time, we could instead look at, for instance, the most recent value of that measurement, or the maximum value ever taken of that measurement across a time period, etc. Once again, choosing between these options could be done using existing clinical knowledge or learned automatically. We provide specific example approaches of how to discretize or summarize features over time in Appendix~\ref{appendix:Other-Preprocessing}, including taking advantage of recently developed machine learning methods. Importantly, our proposed framework accommodates any of these feature preprocessing strategies. We defer studying the effect of using different feature preprocessing strategies to
future work.} Viewing $X$ as an $n$-by-$d$ matrix, the $i$-th row of $X$ (denoted by $X_i$) can be thought of as the feature vector for the $i$-th subject. Importantly, whether death has occurred is not one of the $d$ historical events tracked by the matrix $X$ since we will be predicting time until death.

As for the training label for the $i$-th subject, we have two recordings: indicator $\delta_i \in \{0,1\}$ specifies whether death occurred for the $i$-th subject, and observed time $Y_i \in [0,\infty)$ is the $i$-th subject's ``survival time'' (time until death) if $\delta_i=1$ or the ``censoring time'' if $\delta_i=0$. The idea is that when we stop collecting training data, some subjects are still alive. The $i$-th subject still being alive corresponds to $\delta_i=0$ with a true survival time that is unknown (``censored''); instead, we know that the subject's survival time is at least the censoring time.

\subsection{Topic Modeling}
\label{sec:topic-modeling}
Representing subjects using the matrix $X$ above corresponds to topic modeling. Developed originally to analyze text \citep{blei2003latent}, classically, a topic model represents each text document (in our case, each text document is a subject/patient) by raw counts of how many times $d$ different ``words'' appear in the document (in our case, each word is a binary indicator for whether a past clinical event occurred). These raw counts are stored as the feature vector $X_i$ described previously. A topic model transforms the $i$-th subject's feature vector $X_i$ into a topic weight vector $W_i\in\mathbb{R}^k$, where $W_{i,g}$ measures how much of topic $g\in\{1,2,\dots,k\}$ is present in the $i$-th subject. A common assumption is that the $i$-th subject's feature vector $W_i$ forms a probability distribution, i.e., the $W_{i,g}\ge0$ for all words $g$ and $\sum_{g=1}^k W_{i,g}=1$.  In the context of text documents, examples of topics include ``sports'', ``finance'', and ``movies'', so that a text document could be partially about both sports and finance but not movies, etc. In our case, topics could correspond, for example, to different patient age groups or having a specific severe illness. The goal is to automatically learn these topics.

As a concrete example of a topic model, we review the LDA model by \citet{blei2003latent}. LDA assumes the topic weight vectors $W_i$'s to be generated i.i.d.~from a $k$-dimensional Dirichlet distribution. Next, to relate feature vector $X_i$ to its topic weight vector $W_i$, let $\overline{X}_{i,u}$ denote the fraction of times a word appears for a specific subject, meaning that $\overline{X}_{i,u} = X_{i,u} / \big( \sum_{v=1}^d X_{i,v} \big)$. Then LDA assumes the factorization
\begin{equation}
\overline{X}_{i,u} = \sum_{g=1}^k W_{i,g} A_{g,u}
\label{eq:LDA-factorization}
\end{equation}
for a ``topic-word'' matrix $A\in\mathbb{R}^{k\times d}$, where each row of $A$ is a distribution over the $d$ vocabulary words; rows of $A$ are assumed to be sampled i.i.d.~from a $d$-dimensional Dirichlet distribution. Importantly, the different rows of $A$ correspond to the different topics. Ideally each topic reveals words (or in our usage, historical clinical events) that are considered related or that tend to co-occur. A standard approach is, for example, to examine the most probable words per topic (i.e., identify the words with the highest values per row of~$A$). We remark that equation~\eqref{eq:LDA-factorization} is commonly written compactly as the nonnegative matrix factorization $\overline{X} = W A$, where the matrix $W$ has rows given by the different subjects' topic weight vectors~$W_i$'s.

Given matrix~$X$, LDA estimates the matrices~$W$ and $A$ (along with the parameters of the two Dirichlet distributions that generate rows of~$W$ and~$A$) using variational inference (as done in the original paper by \citet{blei2003latent}) or Gibbs sampling \citep{porteous2008fast}. Recently, \citet{srivastava2017autoencoding} showed how to approximate LDA in a neural net framework so that off-the-shelf neural net optimizers such as Adam \citep{kingma2015adam} can then be used to learn the model.

\paragraph{Prerequisites on the topic model for use with our framework}
Our proposed strategy for combining topic modeling with survival analysis can use any topic model with a neural net formulation that can output an estimate $\widehat{W}$ of the topic weight matrix $W$ stated above. We shall feed $\widehat{W}$ as input to a survival model. We remark that our approach technically does not require the rows of~$W$ to be probability distributions, although as we show later, constraining~$W$ to be nonnegative can ease interpretation of the survival model used.

Aside from LDA, examples of neural topic models that can be used in our survival-supervised topic modeling framework include correlated topic models \citep{lafferty2006correlated}, supervised LDA \citep{mcauliffe2008supervised}, SAGE \citep{eisenstein2011sparse}, ProdLDA \citep{srivastava2017autoencoding}, and the Embedded Topic Model \citep{dieng2019topic}. As there are many neural topic models at this point, we refer the interested reader to the survey by \citet{zhao2021topic}.

\subsection{Survival Analysis}
\label{sec:survival-analysis}
Many standard topic models, including LDA, do not solve a prediction task. To predict time-to-event outcomes, we turn to survival analysis models. In this section, we review some key concepts from survival analysis. More details can be found in standard textbooks (e.g., \citealt{kalbfleisch2002statistical,klein2006survival}). At the end of this section, we state what our approach to combining topic and survival models requires of the survival model used.

Suppose we take the $i$-th subject's feature vector to be $W_i\in\mathbb{R}^k$ instead of $X_i$. As this notation suggests, when we combine topic and survival models, $W_i$ corresponds to the \mbox{$i$-th} subject's topic weight vector; this strategy for combining topic and survival models was first done by \citet{dawson2012survival}, who extended the original supervised LDA formulation by \citet{mcauliffe2008supervised}. We treat the training data to the survival model as $(W_1,Y_1,\delta_1)$, $\dots$, $(W_n,Y_n,\delta_n)$. Thus, the survival model does not get direct access to the ``raw'' feature vectors $X_i$'s. Instead, it only gets information about the raw feature vectors through the topic weight vectors~$W_i$'s.

\paragraph{The prediction task}
A standard survival analysis prediction task can be stated as using the training data $(W_1,Y_1,\delta_1),\dots,(W_n,Y_n,\delta_n)$ to estimate, for any test subject with feature vector $w\in\mathbb{R}^k$, the subject-specific survival function
\[
S(t|w)=\mathbb{P}(\text{subject survives beyond time }t~|~\text{subject's feature vector is }w).
\]
As with standard classification and regression settings, the training and test data are assumed to be i.i.d.~samples from the same underlying distribution.

In survival analysis literature, often the prediction task is instead stated as estimating a transformed version of $S(\cdot|w)$ called the \emph{hazard function}. Formally, let $W_0$ and $T_0$ be continuous random variables corresponding to the test subject's feature vector and the test subject's true survival time. We denote the cumulative distribution function (CDF) of $T_0$ given $W_0$ by $F(t|w)={\mathbb{P}(T_0\le t|W_0=w)}$, and the probability density function (PDF) of this distribution by $f(t|w)=\frac{\partial}{\partial t}F(t|w)$. The survival function is precisely $S(t|w)=1-F(t|w)$. The hazard function is
\begin{equation}
h(t|\weights)
:=-\frac{\partial}{\partial t}\log S(t|\weights)
= \frac{-\frac{\partial}{\partial t} S(t|w)}{S(t|\weights)}
= \frac{-\frac{\partial}{\partial t} [1-F(t|\weights)]}{S(t|\weights)}
= \frac{f(t|\weights)}{S(t|\weights)}, \label{eq:hazard}
\end{equation}
which (from the right-most expression) is the instantaneous rate of death at time $t$ divided by the probability of surviving up to time~$t$, all conditioned on the feature vector being $\weights$. Given how the hazard function is defined, knowing $S(\cdot|w)$ means that we know $h(\cdot|w)$ and vice versa (i.e., if we know $h(\cdot|w)$, then $S(t|w)=\exp(-\int_0^t h(\tau|w)d\tau)$). Naturally, survival models differ in the assumptions they place on the underlying survival function~$S(\cdot|w)$.

The technical challenge in estimating $S(\cdot|w)$ from training data is that in general, we do not observe the survival times for all of the training subjects: the observed times~$Y_i$'s are equal to survival times only for subjects who have~$\delta_i=1$; all other~$Y_i$ values are censoring times. We assume that the $i$-th training subject has survival time $T_i$ and censoring time $C_i$ that are conditionally independent given feature vector $W_i$, and if the survival time occurs before censoring ($T_i\le C_i$), then $Y_i=T_i$ and $\delta_i=1$; otherwise $Y_i=C_i$ and $\delta_i=0$. This setup is referred to as \textit{random censoring}.

\paragraph{Measuring survival prediction accuracy}
Although the prediction task can be described as estimating the survival function $S(\cdot|w)$ (or a variant of it such as the hazard function), when it comes to evaluating accuracy, we do not know the true function $S(\cdot|w)$ even in the training data. A number of evaluation metrics have been devised, for which we use the time-dependent concordance index $C^{\textsf{td}}$ by \citet{antolini2005time}. Roughly, $C^{\textsf{td}}$ measures the fraction of pairs of subjects correctly ordered by a survival model (based on estimated subject-specific survival functions) among pairs of subjects that can be unambiguously ordered. Thus, $C^{\textsf{td}}$ scores are fractions between~0 and~1, and the highest accuracy corresponds to a value of~1.

\paragraph{Prerequisites on the survival model for use with our framework}
Our neural survival-supervised topic modeling framework requires that the survival model used can be learned by (sub)gradient descent using standard neural net optimizers. We will need to backpropagate through both the survival and topic models, which are linked via the topic weight matrix $W$ (estimated by the topic model and treated as the input ``feature vectors'' by the survival model). Numerous survival models satisfy the criterion above of being learnable via (sub)gradient descent including the classical Cox proportional hazards model \citep{cox1972regression} and accelerated failure time (AFT) models (e.g., \citealt{cox1972regression,prentice1978linear}). We state the modeling assumptions of these models next along with their differentiable loss functions and how to construct an estimate $\widehat{S}(\cdot|w)$ for the subject-specific survival function $S(\cdot|w)$ after minimizing each model's loss function.

\subsubsection{Example: Cox Proportional Hazards}
\label{sec:cox}
The Cox model assumes that the hazard function has the form
\begin{equation}
h(t|\weights)
= h_0(t)e^{\beta^{\top}\weights}
\qquad\text{for }t\ge0,~w\in\mathbb{R}^k,
\label{eq:proportional-hazards}
\end{equation}
where the two parameters are the baseline hazard function $h_0:[0,\infty)\rightarrow[0,\infty)$, and the vector of regression coefficients $\beta\in\mathbb{R}^k$. Under random censoring (and actually more general censoring models), we can estimate $\beta$ without knowing~$h_0$ via maximizing a profile likelihood, which is equivalent to minimizing the differentiable loss function
\begin{equation}
L_{\textsf{Cox}}(\beta|W)
=-\frac{1}{n}
  \sum_{i=1}^{n}
    \delta_i
    \Big[\beta^{\top}W_{i}
         -\log\sum_{j=1\text{ s.t.~}Y_j \ge Y_i}^{n}\exp(\beta^{\top}W_j)\Big].
\label{eq:cox-loss}
\end{equation}
After computing parameter estimate $\widehat{\beta}$ by minimizing $L_{\textsf{Cox}}(\beta)$, we can estimate survival functions $S(\cdot|w)$ via the following approach by \citet{breslow1972discussion}. Denote the unique times of death in the training data by $t_1,t_2,\dots,t_m$. Let $d_i$ be the number of deaths at time $t_i$. We first compute the so-called hazard function $\widehat{h}_i:=d_i/(\sum_{j=1\text{ s.t.~}Y_j\ge Y_i}^n e^{\widehat{\beta}^{\top}W_j})$ at each time index $i=1,2,\dots,m$. Next, we form the ``baseline'' survival function $\widehat{S}_{0}(t):=\exp(-\sum_{i=1\text{ s.t.~}t_i \le t}^{m}\widehat{h}_i)$. Finally, subject-specific survival functions are estimated to be powers of the baseline survival function: $\widehat{S}(t|w):=[\widehat{S}_0(t)]^{\exp(\widehat{\beta}^{~\!\top}w)}.$

Importantly, under the Cox model, whether a subject with feature vector~$w$ is predicted to have overall higher or lower survival probabilities across time is determined by the inner product $\widehat{\beta}^\top w = \sum_{g=1}^k \widehat{\beta}_g w_g$. When this inner product is larger, then $\widehat{S}(t|w)=[\widehat{S}_0(t)]^{\exp(\widehat{\beta}^{~\!\top}w)}$ is smaller across time. Recall that we shall take $w$ to be a nonnegative topic weight vector, so the $g$-th topic being present for a subject means that $w_g>0$. Note that the $g$-th topic's contribution to the inner product $\widehat{\beta}^\top w$ is precisely $\widehat{\beta}_g w_g$. Thus, the $g$-th topic having a larger $\widehat{\beta}_g$ coefficient means that the topic is associated with \emph{lower} survival functions/probabilities, and thus \emph{lower} mean (or median) survival times.\footnote{Note that the area under the survival function $\int_0^\infty S(t|w)dt$ is precisely the mean survival time for a subject with feature vector~$w$. The time $t$ for which $S(t|w)$ crosses 1/2 is a median survival time for feature vector~$w$. Thus, when the survival function decreases across all of time (except at time $t=0$, where it is 1), then the mean and median survival times decrease.} By ranking topics based on their $\widehat{\beta}_g$ values, we can thus get a sense of which topics are associated with lower vs higher survival times.

For the above loss $L_{\textsf{Cox}}(\beta)$, we remark that one can regularize the Cox regression coefficients~$\beta$. For example, adding a lasso, ridge, or more generally elastic-net penalty on~$\beta$ leads to the loss minimized by \citet{simon2011regularization}. Adding this regularization does not change how the hazard and survival functions are estimated once we have an estimate $\widehat{\beta}$ of $\beta$. Standard neural net optimizers can accommodate such a regularization term.

\subsubsection{Example: Accelerated Failure Time Models}
\label{sec:aft}
As another example of a survival model that our neural survival-supervised topic modeling framework can use, consider the log-logistic AFT model that assumes each subject's (possibly unobserved) survival time $T_i$ has the form
\begin{equation}
\log T_i = \mu + \theta^\top W_i + \sigma \varepsilon_i,
\label{eq:AFT}
\end{equation}
where $\mu\in\mathbb{R}$, $\theta\in\mathbb{R}^k$, and $\sigma>0$ are model parameters, and the noise variables~$\varepsilon_i$'s are i.i.d.~standard logistic, i.e., $\varepsilon_i$ has PDF $f_{\varepsilon}(s)=e^s/(1+e^s)^2$ and CDF $F_{\varepsilon}(s)=1/(1+e^{-s})$. Thus, $T_i$ given $W_i$ is distributed as a log-logistic distribution and, in particular, the underlying survival function $S(\cdot|W_i)$ has a closed-form expression:
\begin{equation}
S(t|W_i)
= \frac{1}{1 + t^{1/\sigma} \exp\{-(\mu + \theta^\top W_i)/\sigma\}}
\qquad\text{for }t \ge 0.
\label{eq:loglogistic-AFT-surv-function}
\end{equation}
Under random censoring, maximum likelihood estimation for $\mu$, $\theta$, and $\sigma$ is equivalent to minimizing the differentiable loss function
\begin{align}
&L_{\textsf{AFT}}(\theta,\mu,\sigma|W) \nonumber \\
&\quad:= -\frac{1}{n}\sum_{i=1}^{n}\big\{\delta_{i}\log f_{\varepsilon}(z_i)-\delta_{i}\log\sigma+(1-\delta_{i})\log \big(1-F_{\varepsilon}(z_i)\big)\big\},
\label{eq:aft-loss}
\end{align}
where $z_i=(\log Y_i -\mu-\theta^{\top}W_i)/\sigma$. Hence, after minimizing the loss function $L_{\textsf{AFT}}(\theta,\mu,\sigma|W)$, we have estimates $\widehat{\theta}$, $\widehat{\mu}$, and $\widehat{\sigma}$ for $\theta$, $\mu$, and $\sigma$ respectively. We can plug these estimates into equation~\eqref{eq:loglogistic-AFT-surv-function} to come up with an estimate $\widehat{S}(\cdot|w)$ for any feature vector $w$.

Interpretation of the log-logistic AFT model is similar to that of the Cox model. As we take the feature vector $w$ to be a topic weight vector with nonnegative values, once again whether the predicted survival function has higher or lower probabilities is determined by an inner product, this time $\widehat{\theta}^\top w$. However, unlike in the Cox model, where the $g$-th topic having larger Cox regression coefficient~$\widehat{\beta}_g$ means that the $g$-th topic is associated with \emph{shorter} mean/median survival times, for the above AFT model, having larger regression coefficient $\widehat{\theta}_g$ means that the $g$-th topic is associated with \emph{longer} mean/median survival times.\footnote{Under the log-logistic AFT model, the median survival time for a subject with feature vector $w$ is $\exp(\mu+\theta^\top w)$. The mean survival time exists only if $\sigma < 1$ for which it is given by $\frac{\pi\sigma\exp(\mu+\theta^\top w)}{\sin (\pi\sigma)}$.}

Other AFT models are also possible where, for example, $T_i$ given $W_i$ has a log-normal, Weibull, gamma, generalized gamma, or inverse-Gaussian distribution instead of a log-logistic distribution. These different models arise from changing the distribution of the i.i.d.~noise terms $\varepsilon_i$'s in equation~\eqref{eq:AFT}. Moreover, just as with the Cox model, we could introduce regularization.

As stated previously, in this paper we use the time-dependent concordance index accuracy metric, which is based on ranking pairs of subjects. As such, using a ranking-based regularization term when learning a survival model tends to yield higher c-index values, which has been previously reported by other researchers (e.g., \citealt{chapfuwa2018adversarial,lee2018deephit,kvamme2019time}).
Accounting for these previous researchers' findings, in our experiments later when we use an AFT model, we use the same regularization strategy as \citet{chapfuwa2018adversarial} by adding the ranking loss by \citet{raykar2007ranking}:
\begin{equation}
L_{\textsf{ranking}}(\theta)
= - 1 +
  \frac{1}{|\mathcal{E}|}
  \sum_{(i,j)\in\mathcal{E}}
      \log_2\big( 1+e^{\theta^{\top} (W_i - W_j)} \big),
\label{eq:steck}
\end{equation}
where $\mathcal{E}$ consists of pairs of subjects $(i,j)$ such that $\delta_i=1$ (death is observed for the $i$-th training subject) and moreover $Y_j>Y_i$ (the observed time for the $j$-th training subject is higher than that of the $i$-th subject). \citet{raykar2007ranking} show that $-L_{\textsf{ranking}}(\theta)$ is a lower bound on a variant of concordance index; thus, minimizing $L_{\textsf{ranking}}(\theta)$ aims to maximize concordance index.
Note that the Cox model does not need a ranking regularizer since it already approximately maximizes concordance index \citep{raykar2007ranking}.

Importantly, in how we combine neural topic models with survival analysis, for the resulting overall model to be readily interpretable, choosing a simple interpretable survival model is crucial, as we have illustrated with the above Cox and log-logistic AFT examples. Thus, although our approach is indeed compatible with survival models given by deep neural net extensions of Cox and AFT models (e.g., \citealt{faraggi1995neural,katzman2018deepsurv,chapfuwa2018adversarial,kvamme2019time,kvamme2019continuous}) that can be more accurate at time-to-event predictions than classical non-neural-net methods and that can learn highly nonlinear functions of the input feature vector, these deep survival models are typically difficult to interpret.

\section{Neural Survival-Supervised Topic Models}
\label{sec:survival-scholar}

We now present our proposed neural survival-supervised topic modeling framework that can combine any neural topic model and any survival model meeting the prerequisites stated in Sections~\ref{sec:topic-modeling} and~\ref{sec:survival-analysis}. For ease of exposition, we first explain how to combine LDA with the Cox proportional hazards model, similar to what is done by \citet{dawson2012survival} except we do this combination in a neural net framework. To show the flexibility of our framework, we explain how to combine LDA with the log-logistic AFT model, and how to replace LDA with the SAGE topic model.

\subsection{A Neural Formulation of the LDA/Cox Combination}

We first need a neural net formulation of LDA. We can use the \textsc{scholar} framework by \citet{card2018neural}. Card et al.~do not explicitly consider survival analysis in their setup although they mention that predicting different kinds of real-valued outputs can be incorporated by using different label networks. We use their same setup and have the final label network perform survival analysis. We give an overview of \textsc{scholar} before explaining our choice of label network. Note that for clarity of presentation, we present a slightly simplified version of \textsc{scholar}.

The \textsc{scholar} framework specifies a generative model for the data, including how each individual word in each subject is generated. In particular, recall that $X_{i,u}$ denotes the number of times the word $u\in\{1,2,\dots,d\}$ appears for the $i$-th subject. Let $v_i$ denote the number of words for the $i$-th subject, i.e., $v_i=\sum_{u=1}^d X_{i,u}$. We now define the random variable $\psi_{i,\ell}\in\{1,2,\dots,d\}$ to be what the $\ell$-th word for the $i$-th subject is (for $i=1,2,\dots,n$ and $\ell=1,2,\dots,v_i$). Then the generative process for \textsc{scholar} with $k$ topics is as follows, stated for the $i$-th subject:
\begin{enumerate}
\item Generate the $i$-th subject's topic distribution:
\begin{itemize}
\item[(a)] Sample $\widetilde{W}_i\sim\mathcal{N}(\bm{\mu}_0,\text{diag}(\bm{\sigma}_0^2))$, where $\bm{\mu}_0\in\mathbb{R}^k$ and $\bm{\sigma}_0^2\in[0,\infty)^k$ are user-specified, and $\text{diag}(\cdot)$ constructs a diagonal matrix from a vector.
\item[(b)] Set the $i$-th subject's topic weights vector to be $W_i = \text{softmax}(\widetilde{W}_i)$.
\end{itemize}
\item Generate the $i$-th subject's words:
\begin{itemize}
\item[(a)] Compute the $i$-th subject's word distribution $\phi_i = f_{\textsf{word}}(W_i)$, where $f_{\textsf{word}}$ is a generator network.
\item[(b)] For word $\ell=1,2,\dots,v_i$: Sample $\psi_{i,\ell}\sim\text{Multinomial}(\phi_i)$.
\end{itemize}
\item Generate the $i$-th subject's output label:

Sample $Y_i$ from a distribution parameterized by label network $f_{\textsf{label}}(W_i)$.
\end{enumerate}
Different choices for the parameters $\bm{\mu}_0, \bm{\sigma}_0^2$, $f_{\textsf{word}}$, and $f_{\textsf{label}}$ lead to different topic models.
To approximate LDA where topic distributions are sampled from a symmetric Dirichlet distribution with parameter $\alpha>0$, we set $\bm{\mu}_0$ to be the all zeros vector, $\bm{\sigma}_0^2$ to have all entries equal to $(k-1)/(\alpha k)$, and $f_{\textsf{word}}(\weights) = \weights^\top A$, where $A\in\mathbb{R}^{k\times d}$ has a Dirichlet prior per row; in fact the matrix $A$ is the same as the one in equation~\eqref{eq:LDA-factorization}. Standard LDA is unsupervised so step 3 of the above generative process would be omitted. In terms of implementation, we set the $g$-th row of $A$ to be $A_g=\text{softmax}(H_g)$ for an unconstrained matrix $H\in\mathbb{R}^{k\times d}$, and for simplicity, we assume the prior on each row of $A$ to be uniform (a special case of a Dirichlet prior).

\subsubsection{Learning Topic Model Parameters}

The topic model parameters are learned via amortized variational inference \citep{kingma2013auto,rezende2014stochastic}. We summarize this procedure for the above unsupervised LDA neural net approximation including stating the loss function. For the derivation of this procedure and loss function, see Section~3.2 of \citet{card2018neural}.

For the $i$-th subject, we keep track of a distribution $q_i:=\mathcal{N}(\bm{\mu}_i,\text{diag}(\bm{\sigma}_i^2))$, where $\bm{\mu}_i\in\mathbb{R}^k$ and $\bm{\sigma}_i^2\in[0,\infty)^k$ will be defined shortly. Distribution $q_i$ approximates the posterior of unnormalized topic weights~$\widetilde{W}_i$ given the observed words $\psi_i:=(\psi_{i,1},\psi_{i,2},\dots,\psi_{i,v_i})$. We introduce a multilayer perceptron $f_e:\mathbb{R}^d\rightarrow\mathbb{R}^{d'}$ that takes as input $X_i$ (the word counts for the $i$-th subject) and outputs an embedding $\bm{\pi}_i=f_e(X_i)$, where the embedding dimension $d'$ is user-specified. Then we set
\begin{align}
\bm{\mu}_i &= \mathbf{W}_{\mu}\bm{\pi}_i + \bm{b}_{\mu}, \\
\log(\bm{\sigma}_i^2) &= \mathbf{W}_{\sigma}\bm{\pi}_i + \bm{b}_{\sigma}.
\end{align}%
The variables $\mathbf{W}_{\mu}\in\mathbb{R}^{d'\times k}$, $\bm{b}_{\mu}\in\mathbb{R}^{k}$, $\mathbf{W}_{\sigma}\in\mathbb{R}^{d'\times k}$, and $\bm{b}_{\sigma}\in\mathbb{R}^{k}$ are parameters. In the latter equation, $\log$ is applied element-wise. In summary, the model parameters we aim to learn are $\mathbf{W}_{\mu}$, $\bm{b}_{\mu}$, $\mathbf{W}_{\sigma}$, and $\bm{b}_{\sigma}$, the parameters for the multilayer perceptron $f_e$, and finally the matrix $H$ (recall that for LDA, we set  $f_{\textsf{word}}(\weights) = \weights^\top A$ with $A_g=\text{softmax}(H_g)$ in step~2 of the \textsc{scholar} generative process). We collectively refer to all the parameters as $\bm{\Theta}_{\textsf{LDA}}$. Meanwhile, the number of topics $k$, constant $\alpha>0$ (used in the Dirichlet prior for unnormalized topic weights), and the neural architecture of $f_e$ are hyperparameters that are user-specified.

As is standard now in amortized variational inference, the loss function is randomly computed given parameters $\bm{\Theta}_{\textsf{LDA}}$; hyperparameters and the input raw counts matrix $X$ are treated as fixed. For the $i$-th subject, we sample an unnormalized topic weight vector $\widetilde{W}_i^{(\mathsf{s})}\sim q_i$. Then following steps~1(b) and~2(a) of the \textsc{scholar} generative process, we compute the topic weight vector $W_i^{(\mathsf{s})}=\text{softmax}(\widetilde{W}_i^{(\mathsf{s})})$ and word distribution $\zeta_i^{(\mathsf{s})}:=W_i^{(\mathsf{s})\top}A\in[0,1]^d$. We repeat this across all subjects~$i$. Then the loss function minimized by \textsc{scholar} for LDA is
\begin{align}
\widetilde{L}_{\textsf{LDA}}(\bm{\Theta}_{\textsf{LDA}})
&=-\frac{1}{n}
   \sum_{i=1}^n
   \bigg[
   \overbrace{
     \sum_{v=1}^d X_{i,v}
       \log(\zeta_{i,v}^{(\mathsf{s})})
     }^{\!\!\!\!\!\!\!\!\!\!\!\!\!\!\!\!\text{log likelihood of observed words}\!\!\!\!\!\!\!\!\!\!\!\!\!\!\!\!} \nonumber \\
&\qquad\qquad\quad\;
       -
       \underbrace{
         \frac{1}{2}
           \sum_{g=1}^k
             \Big(
               \frac{\bm{\sigma}_{i,g}^2 + \bm{\mu}_{i,g}^2}
                    {(k-1)/(\alpha k)}
               -
               k
               +
               \log \frac{(k-1)/(\alpha k)}{\bm{\sigma}_{i,g}^2}
             \Big)
       }_{\text{KL divergence between }q_i\text{ and true posterior}}
   \bigg].
\label{eq:neural-LDA-loss}
\end{align}
When we apply this framework to clinical data, one practical issue is that some subjects have dramatically more historical clinical measurements than other subjects. For example, in one dataset in our experiments, one subject has a total of 59824 measurements (note that the same ``word''/past historical clinical event could occur multiple times) whereas there is another subject who has a total of 3 measurements! When there is such heterogeneity in how many words are present per ``document''/subject, the subjects with a very large number of historical clinical measurements will dominate the entire loss function above. To prevent this behavior, for all datasets, we replace the raw word counts $X$ with its normalized version $\overline{X}$ stated in Section~\ref{sec:topic-modeling} ($\overline{X}$ is obtained by taking $X$ and dividing each row by the sum of the row), which effectively weights every subject equally (despite subjects possibly having varying amounts of measurements present).\footnote{Other approaches are possible for weighting different subjects. For instance, instead of using the row-normalized matrix $\overline{X}$ or the raw counts matrix $X$, we could interpolate between these two choices by using $\overline{X}_{i,u}^{(\xi)}:= X_{i,u}/(\sum_{v=1}^d X_{i,v})^\xi$, where $\xi\in[0,1]$ is a user-specified hyperparameter (setting $\xi=1$ corresponds to using $\overline{X}_{i,u}$, whereas setting $\xi=0$ corresponds to using the raw count $X_{i,u}$). For simplicity, we simply use $\overline{X}$ in our experiments later.} Thus, the loss function we minimize is instead
\begin{align}
L_{\textsf{LDA}}(\bm{\Theta}_{\textsf{LDA}})
&=-\frac{1}{n}
   \sum_{i=1}^n
   \bigg[
     \sum_{v=1}^d \overline{X}_{i,v}
       \log(\zeta_{i,v}^{(\mathsf{s})})
     \nonumber \\
&\qquad\qquad\quad\;
       -
         \frac{1}{2}
           \sum_{g=1}^k
             \Big(
               \frac{\bm{\sigma}_{i,g}^2 + \bm{\mu}_{i,g}^2}
                    {(k-1)/(\alpha k)}
               -
               k
               +
               \log \frac{(k-1)/(\alpha k)}{\bm{\sigma}_{i,g}^2}
             \Big)
   \bigg].
\label{eq:our-LDA-loss}
\end{align}
We can minimize this loss with respect to $\bm{\Theta}_{\textsf{LDA}}$ using standard neural net optimizers as well as train in minibatches to scale to large datasets. Empirically, \citet{srivastava2017autoencoding} and \citet{card2018neural} have found that for training neural topic models, training with high momentum and using batch normalization is essential in preventing the topics learned from being the same (the so-called issue of ``mode collapse''); for the interested reader, see the implementation notes in Appendix C of~\citet{card2018neural}.

Recall from Section~\ref{sec:topic-modeling} that we require the neural topic model used in our framework to be able to output estimated topic weight vectors $\widehat{W}_i$'s for the different subjects as these will be used as inputs to the survival model. We could simply set $\widehat{W}_i$ to be the topic weight vector $W_i^{(\mathsf{s})}=\text{softmax}(\widetilde{W}_i^{(\mathsf{s})})$ constructed based on the random unnormalized topic weight vector $\widetilde{W}_i^{(\mathsf{s})}\sim q_i$. Alternatively, rather than only using one sample $\widetilde{W}_i^{(\mathsf{s})}$, we could draw multiple samples $\widetilde{W}_i^{(\mathsf{s}),1},\dots,\widetilde{W}_i^{(\mathsf{s}),\ell}\overset{\text{i.i.d.}}{\sim} q_i$, and output $\widehat{W}_i=\frac{1}{\ell}\sum_{j=1}^\ell\text{softmax}(\widetilde{W}_i^{(\mathsf{s}),j})$.

\subsubsection{Survival Supervision}

To incorporate the Cox survival loss, we set step~3 of the \textsc{scholar} generative process to use $f_{\textsf{label}}(W_i)=\beta^\top W_i$ for parameter vector $\beta\in\mathbb{R}^{k}$, where we explicitly constrain $\beta_k=0$, i.e., how much of the $k$-th topic is present is ignored in the inner product calculation. This is done so that the $k$-th topic acts as a background topic. We remark that $f_{\textsf{label}}(W_i)$ is simple to implement: given $W_i$, we drop the entry corresponding to the $k$-th topic and then feed the result to a standard linear layer with a single output node and no bias term. The weights of this fully-connected layer thus correspond to $(\beta_1,\beta_2,\dots,\beta_{k-1})$. The last coefficient $\beta_k=0$ is not stored.

Note that $\beta$ precisely consists of the Cox regression coefficients in equation~\eqref{eq:proportional-hazards}. Meanwhile, $f_{\textsf{label}}(W_i)$ precisely takes the role of the $\beta^\top W_i$ terms in the Cox loss~\eqref{eq:cox-loss}. Of course, as we do not observe the true topic weight vector $W_i$, we plug in its estimate $\widehat{W}_i$ from the topic model. To summarize, the Cox loss we use with the neural topic model is
\begin{align}
&L_{\textsf{Cox-with-background-topic}}(\beta_1,\dots,\beta_{k-1}|\widehat{W}) \nonumber \\
&\quad=
-\frac{1}{n}
\sum_{i=1}^{n}
  \delta_i
  \Big[f_{\textsf{label}}(\widehat{W}_i)
       - \log\sum_{j=1\text{ s.t.~}Y_j \ge Y_i}^{n}\exp(f_{\textsf{label}}(\widehat{W}_j))\Big],
\label{eq:cox-loss-neural}
\end{align}
where we have left out regression coefficient $\beta_k$ as it is constrained to be 0.

We can now state the overall loss function that we minimize for the neural LDA-Cox model:
\begin{align}
&L_{\textsf{LDA-Cox}}(\bm{\Theta}_{\textsf{LDA}}, \beta_1,\dots,\beta_{k-1}) \nonumber \\
&\quad=
L_{\textsf{LDA}}(\bm{\Theta}_{\textsf{LDA}}) + 
\lambda_{\textsf{survival}} L_{\textsf{Cox-with-background-topic}}(\beta_1,\dots,\beta_{k-1}|\widehat{W}),
\end{align}
where hyperparameter $\lambda_{\textsf{survival}}>0$ weights the importance of the survival loss. We refer to the resulting model as \textsc{scholar lda-cox}.

\subsubsection{Model Interpretation}

For the $g$-th topic learned, we can look at its distribution over words $A_g\in[0,1]^d$ (the $g$-th row of $A$ given in equation~\eqref{eq:LDA-factorization}) and, for instance, rank words by their probability of appearing for topic $g$. The $g$-th topic is also associated with Cox regression coefficient $\beta_g$, where each $\beta_g$ is the parameter from equation~\eqref{eq:cox-loss-neural}. Again, the $k$-th topic is constrained to have Cox regression coefficient $\beta_k=0$. Under the Cox model, $\beta_g$ being larger means that the $g$-th topic is associated with \emph{shorter} mean/median survival times, as discussed in Section~\ref{sec:cox}.

\subsection{Using Other Choices of Topic or Survival Models}

To give a sense of the generality of our proposed framework, we explain how to derive neural survival-supervised topic models corresponding to combining LDA with an AFT model (Section~\ref{sec:lda-aft}) as well as combining the SAGE topic model \citep{eisenstein2011sparse} with either Cox or AFT survival models (Section~\ref{sec:replace-lda-with-sage}).

\subsubsection{LDA/AFT}
\label{sec:lda-aft}

To combine LDA with an AFT survival model, we use the same idea as how we combined LDA with a Cox model. The changes are as follows. First off, in step~3 of the \textsc{scholar} generative process, we now set $f_{\textsf{label}}(W_i) = \theta^\top W_i + \mu$, again constraining the $k$-th regression coefficient $\theta_k = 0$ to correspond to a background topic. Effectively, we are taking the survival time $T_i$ to be of the form $\log T_i = f_{\text{label}}(W_i) + \sigma\varepsilon_i$ in equation~\eqref{eq:aft-loss}, where parameters $\mu$, $\theta$, and~$\sigma$ are the same as described in Section~\ref{sec:aft} except with the new constraint that $\theta_k=0$.

Note that $f_{\textsf{label}}(W_i)$ can be implemented by taking the input $W_i$, dropping the $k$-th topic's weight, and then feeding the result through a standard linear layer with one output node and a bias term. The bias term is precisely $\mu$ and the weight matrix of the linear layer precisely gives $(\theta_1,\theta_2,\dots,\theta_{k-1})$. As the true $W_i$ is unknown, we plug in its estimate $\widehat{W}_i$ from the topic model.

We use the regularized survival loss function
\begin{align}
&L_{\textsf{AFT-with-background-topic}}(\mu,\sigma,\theta_1,\dots,\theta_{k-1}|\widehat{W}) \nonumber \\
&\quad=
-\frac{1}{n}\sum_{i=1}^{n}\big\{\delta_{i}\log f_{\varepsilon}(Z_i)-\delta_{i}\log\sigma+(1-\delta_{i})\log \big(1-F_{\varepsilon}(Z_i)\big)\big\} \nonumber \\
&\quad\quad~ + \lambda_{\textsf{ranking}} L_{\textsf{ranking}}(\theta_1,\dots,\theta_{k-1}),
\label{eq:aft-loss-neural}
\end{align}
where $Z_i = [(\log(Y_i)) - f_{\textsf{label}}(\widehat{W}_i)]/\sigma$, $f_\varepsilon(s)={e^s/(1+e^s)^2}$, $F_\varepsilon(s)={1/(1+e^{-s})}$, and $\lambda_{\textsf{ranking}}\ge0$ is a user-specified hyperparameter, and $L_{\textsf{ranking}}(\theta_1,\dots,\theta_{k-1})$ is the same as in equation~\eqref{eq:steck} except with the constraint $\theta_k=0$. Since parameter~$\sigma$ needs to be strictly positive, we instead have the neural net keep track of $\log\sigma$, which is unconstrained and we initialize with a random sample from $\mathcal{N}(0, 10^{-4})$. The overall loss to be minimized is thus
\begin{align}
&L_{\textsf{LDA-AFT}}(\bm{\Theta}_{\textsf{LDA}},\mu,\sigma,\theta_1,\dots,\theta_{k-1}) \nonumber \\
&\quad=L_{\textsf{LDA}}(\bm{\Theta}_{\textsf{LDA}}) + 
\lambda_{\textsf{survival}} L_{\textsf{AFT-with-background-topic}}(\mu,\sigma,\theta_1,\dots,\theta_{k-1}|\widehat{W}),
\end{align}
for a user-specified hyperparameter $\lambda_{\textsf{survival}}>0$. The rest of neural net training works exactly the same way as in the LDA-Cox combination.

As for model interpretation, just as with the LDA-Cox model, for the \mbox{$g$-th} topic, we can inspect its distribution over words given by the $g$-th row of the matrix $A$. As discussed in Section~\ref{sec:aft}, the $g$-th topic has an associated regression coefficient $\theta_g$ for which larger values mean that the $g$-th topic is associated with longer mean/median survival times.

\subsubsection{Replacing LDA with SAGE}
\label{sec:replace-lda-with-sage}

The above LDA/Cox and LDA/AFT combinations can easily accommodate replacing LDA with a different neural topic model. For example, to replace LDA with SAGE \citep{eisenstein2011sparse}, we make the following changes. First, recall that in step 2(a) of  the \textsc{scholar} generative process, the neural net $f_{\textsf{word}}$ maps an input topic weight vector $w$ to a distribution over $d$ words. For SAGE, we set $f_{\textsf{word}}$ to be
\[
f_{\textsf{word}}(w)
= \text{softmax}(\gamma + w^\top H),
\]
where $\gamma\in\mathbb{R}^d$ and $H\in\mathbb{R}^{k \times d}$ are parameters. Note that in a neural net framework, $f_{\textsf{word}}$ is implemented as a linear layer followed by softmax activation. Specifically, the linear layer has a bias term and maps feature vectors of size $k$ to output vectors of size $d$. The linear layer's weight matrix and bias term correspond to $H$ and $\gamma$, respectively.

The interpretation is as follows: given a subject with topic weight vector~$w$, the $v$-th word (a historical clinical event) occurs with probability proportional to $\exp(\gamma_v + \sum_{g=1}^k w_g H_{g, v})$. In this sense, $\gamma_v$ can be thought of as a background log frequency of the $v$-th word. The $g$-th topic is then represented by the $g$-th row of $H$ and can be thought of as log deviations from the background log frequency vector. Phrased informally, SAGE represents each topic as a deviation from background word frequencies. This representation is convenient in that there often are many ``background'' words that appear in a very large fraction of subjects and are not helpful in distinguishing between the topics. For LDA, these background words would have to be removed either as a preprocessing or as a postprocessing step. SAGE on the other hand inherently accounts for these background words.

For SAGE, to interpret the $g$-th topic, we can rank words the words from largest to smallest deviation from background according to the values in the $g$-th row of $H$. The values are of course not probabilities. For example, for the $g$-th topic, if the $v$-th word has a log deviation value $H_{g,v}=3$, then it means that it occurs $\exp(3)$ times more than word $v$'s background frequency. It is of course possible to have negative log deviation values.

The loss function we use to learn the SAGE topic model is almost the same as for LDA and is given by
\begin{align}
L_{\textsf{SAGE}}(\bm{\Theta}_{\textsf{SAGE}})
&=-\frac{1}{n}
   \sum_{i=1}^n
   \bigg[
     \sum_{v=1}^d \overline{X}_{i,v}
       \log(\zeta_{i,v}^{(\mathsf{s})}) \nonumber \\
&\qquad\qquad\quad\;
       -
       \frac{1}{2}
         \sum_{g=1}^k
           \Big(
             \frac{\bm{\sigma}_{i,g}^2 + \bm{\mu}_{i,g}^2}
                  {(k-1)/(\alpha k)}
             -
             k
             +
             \log \frac{(k-1)/(\alpha k)}{\bm{\sigma}_{i,g}^2}
           \Big)
   \bigg] \nonumber \\
&\quad + \lambda_{\textsf{small-deviation}} \sum_{g=1}^k \sum_{v=1}^d H_{g,v}^2,
\end{align}
where the differences are that: (a) we redefine $\zeta_{i}^{(\mathsf{s})} = {\text{softmax}(\gamma + W_i^{(\textsf{s})\top}H)}$, and (b) we add an $\ell_2$ regularization term on the log deviations, with a user-specified weight $\lambda_{\textsf{small-deviation}}\ge0$. The rest of the setup is the same as for LDA, and we collectively denote the complete set of parameters that we minimize the loss over as $\bm{\Theta}_{\textsf{SAGE}}$. By combining this topic model with the Cox and log-logistic AFT survival models, we obtain \textsc{scholar sage-cox} and \textsc{scholar sage-aft}.

We remark that the original SAGE model actually also uses $\ell_1$ regularization on the log deviations in $H$, but in preliminary experiments, we found that encouraging sparsity yields topic models that are not clinically interpretable. The issue is that in healthcare, often times, a collection of clinical measurements help explain a condition. When these measurements are collinear or have high pairwise correlation, enforcing sparsity would favor just retaining one of these measurements and zeroing out the contributions of the others \citep[Section 2.3]{zou2005regularization}. Consequently, we lose valuable co-occurrence information of related clinical features. For this reason, as well as the previous empirical finding by \citet{card2018neural} that encouraging sparsity results in worse topics learned in terms of other standard topic modeling metrics of perplexity and coherence, we do not encourage sparsity in learning the topic log deviations matrix $H$.

\section{Experiments}
\label{sec:experiments}

\subsection{Data}

We conduct experiments on seven datasets: data on severely ill hospitalized patients from the Study to Understand Prognoses Preferences Outcomes and Risks of Treatment (SUPPORT) \citep{knaus1995support}, which---as suggested by \citet{harrell2015regression}---we split into four datasets corresponding to different disease groups (acute respiratory failure/multiple organ system failure, cancer, coma, COPD/congestive heart failure/cirrhosis); data from breast cancer patients (METABRIC) \citep{curtis2012genomic}; data from patients who received heart transplants in the United Network for Organ Sharing (UNOS);\footnote{We use the UNOS Standard Transplant and Analysis Research data from the Organ Procurement and Transplantation Network as of September 2019, requested at: \url{https://www.unos.org/data/}} and lastly patients with intracerebral hemorrhage (ICH) from the MIMIC-III electronic heath records dataset \citep{johnson2016physionet,johnson2016mimic}. For all except the last dataset, we predict time until death; for the ICH patients, we predict time until discharge from a hospital ICU. Basic characteristics of these datasets are reported in Table~\ref{tab:datasets}. More details on the datasets and on data preproprocessing are in~\ref{appendix:data-preprocessing}. We randomly divide each dataset into a 80\%/20\% train/test split. Our code is publicly available.\footnote{\url{https://github.com/georgehc/survival-topics}}

\begin{table}[t]
\caption{Basic characteristics of the survival datasets used.}
\scriptsize
\centering
{\renewcommand{\arraystretch}{1.2}
\begin{tabular}{ccccc}\toprule
\multirow{2}{*}{Dataset}   & \multirow{2}{*}{Description} & Number of  & Number of & Fraction \\
& & subjects & features & censored \\
\midrule
\multirow{2}{*}{\textsc{support1}} & acute respiratory failure/multiple organ & \multirow{2}{*}{4203} & \multirow{2}{*}{14} & \multirow{2}{*}{35.7\%} \\
& system failure & & & \\
\multirow{2}{*}{\textsc{support2}} & chronic obstructive pulmonary disease/congestive & \multirow{2}{*}{2854} & \multirow{2}{*}{14} & \multirow{2}{*}{39.4\%} \\
& heart failure/cirrhosis & & & \\
\textsc{support3} & cancer & 1413 & 13 & 11.3\% \\
\textsc{support4} & coma & 592 & 14 & 18.8\% \\
\textsc{metabric} & breast cancer & 1981 & 24  & 55.2\% \\
\textsc{unos} & heart transplant & 62644 & 49  & 50.2\% \\
\textsc{mimic-ich} & intracerebral hemorrhage & 961 & 1530 & 23.1\% \\\bottomrule
\end{tabular}
}
\label{tab:datasets}
\end{table}

\subsection{Experimental Setup}

We benchmark \textsc{scholar lda-cox}, \textsc{scholar lda-aft}, \textsc{scholar sage-cox}, and \textsc{scholar sage-aft} against~5 baselines: 2 classical methods (lasso-regularized Cox \citep{simon2011regularization}, and random survival forests (RSF) \citep{ishwaran2008random}), 2 deep learning methods (DeepSurv \citep{katzman2018deepsurv} and DeepHit \citep{lee2018deephit}), and a naive two-stage decoupled LDA/Cox model (fit unsupervised LDA first and then fit a Cox model). For all methods, we hold out 20\% of the training data as a validation set to select hyperparameters. Hyperparameter search grids are included in Appendix~\ref{appendix:hyperparameter-grids}. For evaluating a model's prediction accuracy on the validation set as well as the final test set, we use the time-dependent concordance $C^{\textsf{td}}$ index \citep{antolini2005time}. For every test set $C^{\textsf{td}}$ index reported, we also compute its 95\% confidence interval, which we obtain by taking 1000 bootstrap samples of the test set with replacement, recomputing the $C^{\textsf{td}}$ index per bootstrap sample, and taking the 2.5 and 97.5 percentile values among the $C^{\textsf{td}}$ indices computed.

\begingroup
\setlength{\intextsep}{0pt}

\subsection{Results}

Test set $C^{\textsf{td}}$ indices are reported in Table~\ref{tab:results} with the 95\% bootstrap confidence intervals. The main takeaways are that:
\begin{itemize}

\item Random survival forest is clearly a strong baseline for the datasets considered, often outperforming the deep learning baselines \textsc{deepsurv} and \textsc{deephit}. That said, no single model is consistently the best.

\item The different neural survival-supervised topic models tested have accuracy scores that are often quite similar with each other.

\item The neural survival-supervised topic models often achieve accuracy scores as good as deep neural net baselines. For example, if we ignore the confidence intervals for a moment and go by test set $C^{\textsf{td}}$ index alone, \textsc{scholar lda-cox}'s accuracy scores on \textsc{support2}, \textsc{unos}, and \textsc{mimic-ich} are better than those of \textsc{deepsurv}. Meanwhile, \textsc{scholar lda-cox}'s accuracy scores on \textsc{support3}, \textsc{metabric}, \textsc{unos}, and \textsc{mimic-ich} are better than those of \textsc{deephit}. We remark that these differences though are often small and, especially once we account for the confidence intervals, we would not claim that neural survival-supervised topic models yield more accurate predictions than the deep learning baselines or vice versa. Rather we would say that these different neural net approaches are competitive with each other when it comes to prediction accuracy as measured by $C^{\textsf{td}}$ index.

\item Clearly, the naive approach (\textsc{naive lda-cox}) of fitting an unsupervised topic model first and then separately training a Cox model using the topics learned tends to achieve worse accuracy scores than its supervised counterpart \textsc{scholar lda-cox}.
\end{itemize}

\begin{table}[t]
\caption{Test set $C^{\textsf{td}}$ indices with 95\% bootstrap confidence intervals.\label{tab:results}}
\scriptsize
\centering
{\renewcommand{\arraystretch}{1.2}
\begin{tabular}{cccccccc}\toprule
\multirow{2}{*}{Model} & \multicolumn{7}{c}{Dataset}\\
\cmidrule(lr){2-8}
 & \textsc{support1} & \textsc{support2} & \textsc{support3} & \textsc{support4} & \textsc{metabric} & \textsc{unos} & \textsc{mimic-ich} \\\midrule
\multirow{3}{*}{\textsc{cox}} & 0.631 & 0.555 & \textbf{0.580} & 0.527 & 0.675 & 0.594 & 0.612 \\
& (0.608, & (0.520, & (0.541, & (0.459, & (0.630, & (0.585, & (0.551, \\
& \phantom{(}0.656) & \phantom{(}0.590) & \phantom{(}0.616) & \phantom{(}0.596) & \phantom{(}0.715) & \phantom{(}0.602) & \phantom{(}0.659) \\
\cmidrule[0.01em](l{.75em}r{.75em}){1-8}
\multirow{3}{*}{\textsc{rsf}} & \textbf{0.657} & 0.578 & 0.562 & \textbf{0.550} & \textbf{0.712} & \textbf{0.604} & 0.618 \\
& (0.631, & (0.547, & (0.522, & (0.481, & (0.670, & (0.596, & (0.567, \\
& \phantom{(}0.684) & \phantom{(}0.611) & \phantom{(}0.603) & \phantom{(}0.616) & \phantom{(}0.755) & \phantom{(}0.612) & \phantom{(}0.666) \\
\cmidrule[0.01em](l{.75em}r{.75em}){1-8}
\multirow{3}{*}{\textsc{deepsurv}} & 0.644 & 0.551 & 0.573 & 0.522 & 0.706 & 0.597 & 0.615 \\
& (0.619, & (0.518, & (0.536, & (0.452, & (0.667, & (0.588, & (0.565, \\
& \phantom{(}0.671) & \phantom{(}0.583) & \phantom{(}0.611) & \phantom{(}0.588) & \phantom{(}0.745) & \phantom{(}0.604) & \phantom{(}0.667) \\
\cmidrule[0.01em](l{.75em}r{.75em}){1-8}
\multirow{3}{*}{\textsc{deephit}} & 0.636 & \textbf{0.579} & 0.549 & 0.531 & 0.666 & 0.585 & 0.587 \\
& (0.610, & (0.545, & (0.509, & (0.458, & (0.620, & (0.576, & (0.533, \\
& \phantom{(}0.662) & \phantom{(}0.613) & \phantom{(}0.590) & \phantom{(}0.594) & \phantom{(}0.710) & \phantom{(}0.593) & \phantom{(}0.637) \\
\cmidrule[0.01em](l{.75em}r{.75em}){1-8}
\multirow{3}{*}{\setstackgap{L}{2.05ex}\Centerstack{\textsc{naive}\\\textsc{lda-cox}}} & 0.543 & 0.536 & 0.508 & 0.541 & 0.639 & 0.540 & 0.537 \\
& (0.517, & (0.504, & (0.470, & (0.471, & (0.589, & (0.532, & (0.484, \\
& \phantom{(}0.567) & \phantom{(}0.571) & \phantom{(}0.545) & \phantom{(}0.614) & \phantom{(}0.686) & \phantom{(}0.549) & \phantom{(}0.591) \\
\cmidrule[0.01em](l{.75em}r{.75em}){1-8}
\multirow{3}{*}{\setstackgap{L}{2.05ex}\Centerstack{\textsc{scholar}\\\textsc{lda-cox}}} & 0.636 & 0.559 & 0.569 & 0.510 & 0.696 & 0.600 & \textbf{0.639} \\
& (0.612, & (0.528, & (0.533, & (0.439, & (0.653, & (0.591, & (0.588, \\
& \phantom{(}0.662) & \phantom{(}0.591) & \phantom{(}0.608) & \phantom{(}0.572) & \phantom{(}0.737) & \phantom{(}0.608) & \phantom{(}0.687) \\
\cmidrule[0.01em](l{.75em}r{.75em}){1-8}
\multirow{3}{*}{\setstackgap{L}{2.05ex}\Centerstack{\textsc{scholar}\\\textsc{lda-aft}}} & 0.631 & 0.544 & 0.531 & 0.493 & 0.688 & 0.596 & 0.634 \\
& (0.606, & (0.510, & (0.494, & (0.427, & (0.643, & (0.588, & (0.585, \\
& \phantom{(}0.657) & \phantom{(}0.579) & \phantom{(}0.571) & \phantom{(}0.556) & \phantom{(}0.728) & \phantom{(}0.604) & \phantom{(}0.680) \\
\cmidrule[0.01em](l{.75em}r{.75em}){1-8}
\multirow{3}{*}{\setstackgap{L}{2.05ex}\Centerstack{\textsc{scholar}\\\textsc{sage-cox}}} & 0.633 & 0.522 & 0.561 & 0.516 & 0.708 & 0.603 & 0.629 \\
& (0.607, & (0.488, & (0.526, & (0.442, & (0.669, & (0.595, & (0.579, \\
& \phantom{(}0.657) & \phantom{(}0.557) & \phantom{(}0.598) & \phantom{(}0.591) & \phantom{(}0.746) & \phantom{(}0.611) & \phantom{(}0.677) \\
\cmidrule[0.01em](l{.75em}r{.75em}){1-8}
\multirow{3}{*}{\setstackgap{L}{2.05ex}\Centerstack{\textsc{scholar}\\\textsc{sage-aft}}} & 0.604 & 0.560 & 0.554 & 0.517 & 0.700 & 0.599 & 0.631 \\
& (0.579, & (0.526, & (0.511, & (0.450, & (0.659, & (0.591, & (0.579, \\
& \phantom{(}0.630) & \phantom{(}0.593) & \phantom{(}0.595) & \phantom{(}0.581) & \phantom{(}0.742) & \phantom{(}0.606) & \phantom{(}0.681) \\
\bottomrule
\end{tabular}
}
\end{table}

To supplement our third takeaway above, specifically for \textsc{scholar lda-cox}, we also use bootstrap sampling to compute differences between $C^{\mathsf{td}}$ indices of \textsc{scholar lda-cox} vs different baseline models. Specifically, we repeat the following 1000 times: (a) take a bootstrap sample from the test set, (b) compute the bootstrap sample's predictions using \textsc{scholar lda-cox} as well as a baseline model, (c) compute the $C^{\textsc{td}}$ index of \textsc{scholar lda-cox}'s predictions minus that of the baseline model's predictions. Thus, we have 1000 differences in $C^{\textsc{td}}$ indices, for which we then take the 2.5 and 97.5 percentiles to get a 95\% confidence interval. We report these confidence intervals in Table~\ref{tab:bootstrap-scholar-lda-cox-baseline-comparison}. We find that 0 is in all the confidence intervals for \textsc{scholar lda-cox} vs \textsc{deepsurv} and nearly in all the ones for \textsc{scholar lda-cox} vs \textsc{deephit} (in fact, the only times 0 is not included for \textsc{deephit} is for the \textsc{unos} and \textsc{mimic-ich} datasets, in which \textsc{scholar lda-cox} is more accurate). We omit tables that compare the other neural survival-supervised topic models with various baselines as they follow similar trends. To reiterate, we do not claim that our proposed models outperform the various baselines tested. Instead we claim that they achieve accuracy that is competitive with deep learning baselines. In fact, Tables~\ref{tab:results} and~\ref{tab:bootstrap-scholar-lda-cox-baseline-comparison} suggest that \textsc{scholar lda-cox} is competitive with \textsc{cox} and \textsc{rsf} as well. On the other hand, the \textsc{naive lda-cox} baseline does appear to be significant less accurate than \textsc{scholar lda-cox} for all datasets except \textsc{support2} and \textsc{support4}.

\begin{table}[!t]
\caption{95\% bootstrap confidence intervals for the test set $C^{\textsf{td}}$ index of \textsc{scholar-lda} minus that of various baselines (when this difference is positive, it means that \textsc{scholar-lda} is more accurate than a particular baseline). Note that for \textsc{rsf}, the \mbox{``-0.000''} value actually corresponds to -0.000142. \label{tab:bootstrap-scholar-lda-cox-baseline-comparison}}
\vspace{.5em}
\scriptsize
\centering
{\renewcommand{\arraystretch}{1.2}
\begin{tabular}{cccccccc}\toprule
\multirow{2}{*}[-2.5pt]{Baseline} & \multicolumn{7}{c}{Dataset} \\
\cmidrule(lr){2-8}
 & \textsc{support1} & \textsc{support2} & \textsc{support3} & \textsc{support4} & \textsc{metabric} & \textsc{unos} & \textsc{mimic-ich} \\ \midrule
\multirow{2}{*}{\textsc{cox}}
 & (-0.008, & (-0.020, & (-0.043, & (-0.098, & (-0.015, & (0.002, & (-0.029, \\
 & \phantom{(}0.019) & \phantom{(}0.028) & \phantom{(}0.027) & \phantom{(}0.057) & \phantom{(}0.059) & \phantom{(}0.010) & \phantom{(}0.088) \\
\cmidrule[0.01em](l{.75em}r{.75em}){1-8}
\multirow{2}{*}{\textsc{rsf}}
 & (-0.037, & (-0.050, & (-0.026, & (-0.106, & (-0.041, & (-0.010, & (-0.027, \\
 & \phantom{(}-0.004) & \phantom{(}0.010) & \phantom{(}0.043) & \phantom{(}0.028) & \phantom{(}0.009) & \phantom{(}-0.000) & \phantom{(}0.070) \\
\cmidrule[0.01em](l{.75em}r{.75em}){1-8}
\multirow{2}{*}{\textsc{deepsurv}}
 & (-0.025, & (-0.025, & (-0.039, & (-0.088, & (-0.039, & (-0.002, & (-0.010, \\
 & \phantom{(}0.009) & \phantom{(}0.044) & \phantom{(}0.036) & \phantom{(}0.062) & \phantom{(}0.019) & \phantom{(}0.009) & \phantom{(}0.059) \\
\cmidrule[0.01em](l{.75em}r{.75em}){1-8}
\multirow{2}{*}{\textsc{deephit}}
 & (-0.018, & (-0.056, & (-0.020, & (-0.092, & (-0.006, & (0.007, & (0.010, \\
 & \phantom{(}0.019) & \phantom{(}0.017) & \phantom{(}0.064) & \phantom{(}0.049) & \phantom{(}0.069) & \phantom{(}0.024) & \phantom{(}0.100) \\
\cmidrule[0.01em](l{.75em}r{.75em}){1-8}
\multirow{2}{*}{\Centerstack{\textsc{naive}\\\textsc{lda-cox}}}
 & (0.063, & (-0.010, & (0.020, & (-0.139, & (0.028, & (0.053, & (0.031, \\
 & \phantom{(}0.123) & \phantom{(}0.055) & \phantom{(}0.104) & \phantom{(}0.062) & \phantom{(}0.088) & \phantom{(}0.066) & \phantom{(}0.170) \\
\bottomrule
\end{tabular}
}
\end{table}

\subsection{Interpretability of Baselines}

Importantly, we remark that the deep learning baselines \textsc{deepsurv} and \textsc{deephit} do not produce interpretable models and they were not designed to be interpretable. Random survival forests are also not easily interpretable: while a single decision tree could be interpretable if its depth and number of leaves are not too large, the difficulty in interpreting a learned random survival forest model is that there are many trees (in our experiments, we use 100 trees for each model), and the best-performing models tend to have learned trees that are moderate in size (e.g., a depth of 6 with 64 leaves). Having to look at 100 moderate-sized trees to interpret a single random survival forest model is not that simple, and it is not straightforward teasing apart how features are related without instead using some post hoc explanation approach like SHAP \citep{lundberg2017unified} or TreeExplainer \citep{lundberg2020local}. Of the models evaluated, only the Cox model and the survival-supervised topic models can readily be interpreted. However, as mentioned in Section~\ref{sec:intro}, Cox models do not inherently learn how features relate, and one would have to introduce new features that encode interactions, which becomes impractical when the number of features is large.

\subsection{Interpretability of Neural Survival-Supervised Topic Models}

We next discuss interpretability of neural survival-supervised topic models. As there are many models considered, for ease of exposition, we only present results for \textsc{scholar lda-cox}, for which we provide a complete summary of all topics learned for the seven datasets along with a detailed look at a few datasets. We remark that clinical expertise is required to interpret the topics.

We begin with summaries of the topics learned. Back in Section~\ref{sec:intro}, we already presented one such summary for the \textsc{support3} dataset in Table~\ref{tab:support3-scholar-lda-cox-topic-summary}. The summaries for the rest of the datasets are in Tables \ref{tab:support1-scholar-lda-cox-topic-summary}, \ref{tab:support2-scholar-lda-cox-topic-summary}, \ref{tab:support4-scholar-lda-cox-topic-summary}, \ref{tab:metabric-scholar-lda-cox-topic-summary}, \ref{tab:unos-scholar-lda-cox-topic-summary}, and~\ref{tab:ich-scholar-lda-cox-topic-summary}. For each topic, we state both the Cox $\beta$ regression coefficient as well as the topic interpretation. For all datasets except \textsc{mimic-ich}, larger $\beta$ corresponds to \emph{shorter} mean/median survival time. For \textsc{mimic-ich}, larger $\beta$ corresponds to \emph{shorter} mean/median hospital length of stay. Note that sometimes, spurious topics are found, where a clinical interpretation readily reveals that we could have used a fewer number of topics (although the hyperparameter selection procedure we use that chooses the best model based on validation $C^{\textsf{td}}$ index would not know this). Overall, seeking a clinical interpretation of topics was straightforward. In contrast, when, for example, we presented topics learned using a neural survival-supervised topic model that encouraged sparsity, a clinical expert was unable to determine what the topics meant, with a key problem raised being that the features that are most probable per topic did not appear to be related to each other. We suspect that this has to do with the known issue with lasso regularization where within a group of features that have high pairwise correlation, lasso will arbitrarily choose one of these features and give 0 weight to the others \citep[Section 2.3]{zou2005regularization}.

To obtain the topic interpretations for each dataset, we filter out features that appear in too few or too many patients. Importantly,  following the work of \citet{schofield2017pulling}, we filter features \emph{after} learning a topic model in contrast to doing so \emph{before} learning the model. Schofield et al.~empirically find no advantage in filtering features before learning a topic model compared to doing it afterward. For our purposes, filtering features before learning a topic model presents problems since there are too many possible ways to do this filtering, and it is unclear how these different filtering approaches impact the topics that are learned. \citet{dawson2012survival} for example use a heuristic preprocessing step in how they use \textsc{survLDA} where they cluster subjects based on their survival outcomes and screen out features that are not sufficiently different between the clusters. The problem is that there are far too many choices of how to do this clustering and how to decide what features are sufficiently different even before learning the topic model. By instead filtering features after learning the model, we leave this choice up to the user to specify. The benefits are that there is no need to retrain the model when we try different filters, and moreover, the filtering is fast so it can be adjusted on demand, for example accounting for clinician input. For the results that we show on learned topics by \textsc{scholar lda-cox}, we specifically filter out features that appear in fewer than 2\% of the patients or more than 50\% of the patients. Essentially features that are too rare do not help explain enough of the patient cohort, and features that are too common do not help with interpretation. We tried different thresholds and found ones that appear to work reasonably well across all datasets.

\begin{table}
\caption{Summary of topics learned by \textsc{scholar lda-cox} on the \textsc{support1} (acute respiratory failure, multiple organ system failure) dataset. Higher $\beta$ is associated with shorter survival time.}
\label{tab:support1-scholar-lda-cox-topic-summary}
\centering
\footnotesize
{\renewcommand{\arraystretch}{1.2}
\begin{tabular}{cp{0.8\linewidth}}\toprule
$\beta$ & Topic interpretation\\\midrule
0 & with cancer, metastases, electrolyte abnormalities, vitals\\
$-5.05$ & protective, female, diabetic\\
$-5.43$ & protective, young, no comorbidity\\\bottomrule
\end{tabular}
}
\end{table}

\begin{table}
\caption{Summary of topics learned by \textsc{scholar lda-cox} on the \textsc{support2} (COPD, congestive heart failure, cirrhosis) dataset. Higher $\beta$ is associated with shorter survival time.}
\label{tab:support2-scholar-lda-cox-topic-summary}
\centering
\footnotesize
{\renewcommand{\arraystretch}{1.2}
\begin{tabular}{cp{0.8\linewidth}}\toprule
$\beta$ & Topic interpretation\\\midrule
5.30 & old, comorbid\\
2.72 & middle age, less comorbid, tachycardia\\
0 & Young healthy baseline, tachycardia\\\bottomrule
\end{tabular}
}
\end{table}

\begin{table}
\caption{Summary of topics learned by \textsc{scholar lda-cox} on the \textsc{support4} (coma) dataset. Higher $\beta$ is associated with shorter survival time.}
\label{tab:support4-scholar-lda-cox-topic-summary}
\centering
\footnotesize
{\renewcommand{\arraystretch}{1.2}
\begin{tabular}{cp{0.8\linewidth}}\toprule
$\beta$ & Topic interpretation\\\midrule
0.47 & kidney failure, tachycardia, hypertensive, comorbid\\
0.08 & respiratory distress/MV, infection/inflammation, hypothermic\\
0.01 & hypothermic otherwise normal\\
0 & normal baseline\\
$-0.00011$ & kidney failure, old, infection/inflammation\\
$-0.58$ & healthy\\\bottomrule
\end{tabular}
}
\end{table}

\begin{table}
\caption{Summary of topics learned by \textsc{scholar lda-cox} on the \textsc{metabric} (breast cancer) dataset. Higher $\beta$ is associated with shorter survival time.}
\label{tab:metabric-scholar-lda-cox-topic-summary}
\centering
\footnotesize
{\renewcommand{\arraystretch}{1.2}
\begin{tabular}{cp{0.8\linewidth}}\toprule
$\beta$ & Topic interpretation\\\midrule
1.29 & er- pr- her2+, high mortality, advanced grade\\
0 & similar to 1, focus on group 4 not 1, site 1 not 3\\
$-1.20$ & protective her2\_status1 (-) er- pr-\\
$-1.29$ & protective but high cellularity luma; pr+ er+\\
\cmidrule(lr){2-2}
$-1.37$ & \multirow{2}{*}{these last two topics are both on protective low npi} \\
$-1.38$ & \\\bottomrule
\end{tabular}
}
\end{table}

\begin{table}
\caption{Summary of topics learned by \textsc{scholar lda-cox} on the \textsc{unos} (heart transplant) dataset. Higher $\beta$ is associated with shorter survival time.}
\label{tab:unos-scholar-lda-cox-topic-summary}
\centering
\footnotesize
{\renewcommand{\arraystretch}{1.2}
\begin{tabular}{cp{0.8\linewidth}}\toprule
$\beta$ & Topic interpretation\\\midrule
6.92 & old, old donor, renal failure, with transfusions, liver failure, previous transplant\\
0 & baseline, heart failure, diabetes, with lvad\\
$-1.45$ & panel reactive antibodies, middle age, low ischemic time, inotropes, body measurements (height weight bmi)\\
$-5.04$ & pediatric cases, young, donor with infection\\
\cmidrule(lr){2-2}
$-5.09$ & \multirow{2}{*}{\parbox{0.8\linewidth}{\emph{these last two topics appear to be spurious and are a mix of the topics with $\beta$ coefficients 0 and $-$5.04}}} \\
$-5.17$ & \\\bottomrule
\end{tabular}
}
\end{table}

\begin{table}
\caption{Summary of topics learned by \textsc{scholar lda-cox} on the \textsc{mimic-ich} (intracerebral hemorrhage) dataset. Higher $\beta$ is associated with shorter hospital length of stay.}
\label{tab:ich-scholar-lda-cox-topic-summary}
\centering
\footnotesize
{\renewcommand{\arraystretch}{1.2}
\begin{tabular}{cp{0.8\linewidth}}\toprule
$\beta$ & Topic interpretation\\\midrule
2.08 & relatively healthy, anticoagulated, protective demographic factors \\
1.34 & severe anemia, renal failure, inflammatory profile  \\
1.14 & hematuria, thrombocytopenia \\
0 & negative drug screening \\
$-2.05$ & glycosuria screen, electrolyte abnormalities \\\bottomrule
\end{tabular}
}
\end{table}

In addition to filtering features, we also provide heatmap visualizations.  These heatmaps were presented to a clinician to obtain the summaries in Tables \ref{tab:support3-scholar-lda-cox-topic-summary}, \ref{tab:support1-scholar-lda-cox-topic-summary}, \ref{tab:support2-scholar-lda-cox-topic-summary}, \ref{tab:support4-scholar-lda-cox-topic-summary}, \ref{tab:metabric-scholar-lda-cox-topic-summary}, \ref{tab:unos-scholar-lda-cox-topic-summary}, and \ref{tab:ich-scholar-lda-cox-topic-summary}. In Section~\ref{sec:intro}, we already presented one such heatmap for the \textsc{support3} dataset in Figure~\ref{fig:support3-heatmap}. Heatmaps for the other datasets are shown in Figures~\ref{fig:support1-heatmap}, \ref{fig:support2-heatmap}, \ref{fig:support4-heatmap}, \ref{fig:metabric-heatmap}, \ref{fig:unos-heatmap}, and \ref{fig:ich-heatmap}; note that for the \textsc{unos} and \textsc{mimic-ich} datasets, due to the large number of features, we truncate the heatmap to only show the top $\sim$80 features (since we only display categorical variables as a block of features at once, we do not get to exactly 80). In these heatmaps, the columns index different topics (with Cox $\beta$ regression coefficient displayed per topic; the topics are sorted in decreasing order of $\beta$ coefficient). The rows index different features. The features are sorted based on the maximum word probability across topics (i.e., for the $k$-by-$d$ topic-word matrix $A$, for the $v$-th column/word, we sort based on the score $\max_{g=1,\dots,k} A_{g,v}$). Furthermore, after doing this sorting, we group together features corresponding to the same categorical variable. Note that we only show features that meet the filtering requirements stated previously.

\begin{figure}
\centering
\includegraphics[scale=0.7]{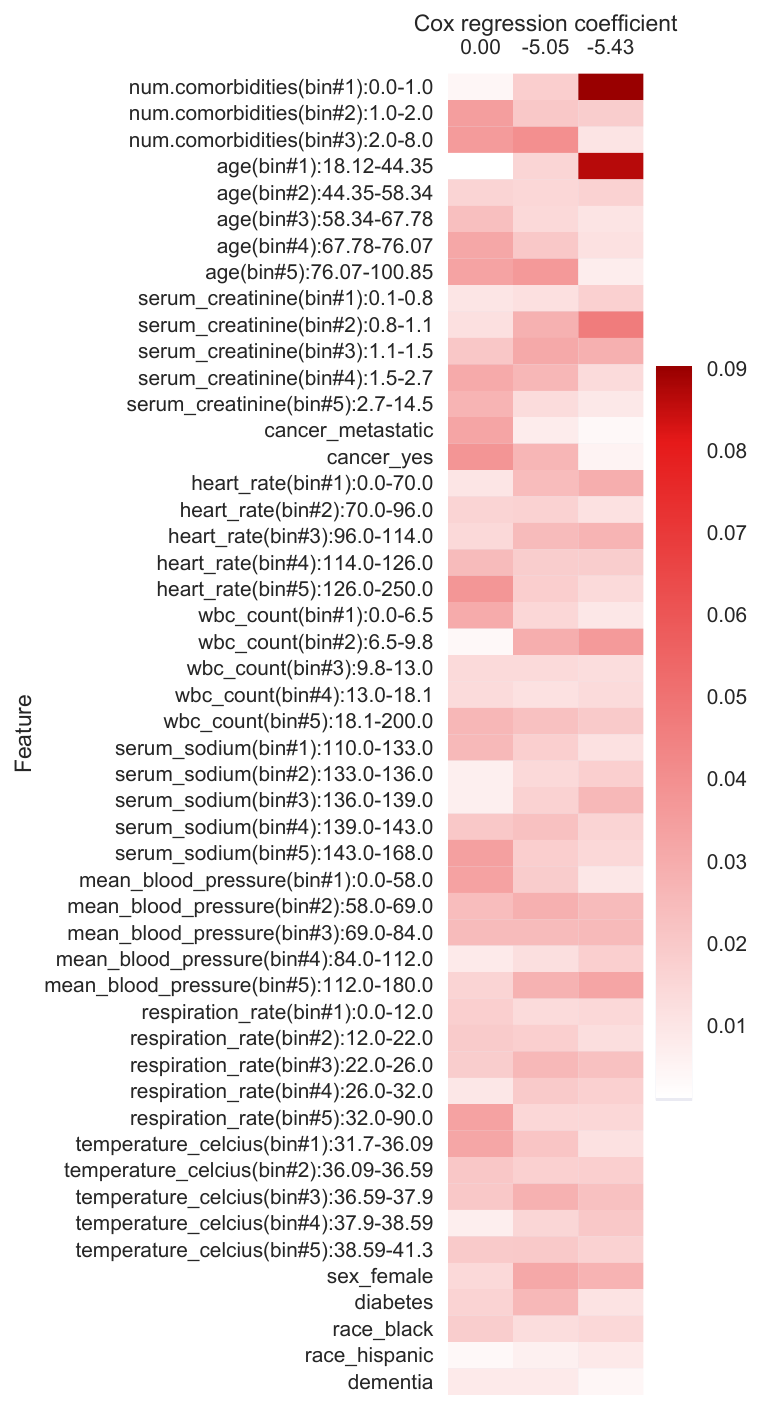}
\caption{Topics learned by \textsc{scholar lda-cox} on the \textsc{support1} (acute respiratory failure/multiple organ system failure) dataset. Columns index topics and rows index features/``words''. The values are probabilities of each feature conditioned on being in a topic.}
\label{fig:support1-heatmap}
\end{figure}

\begin{figure}
\centering
\includegraphics[scale=0.7]{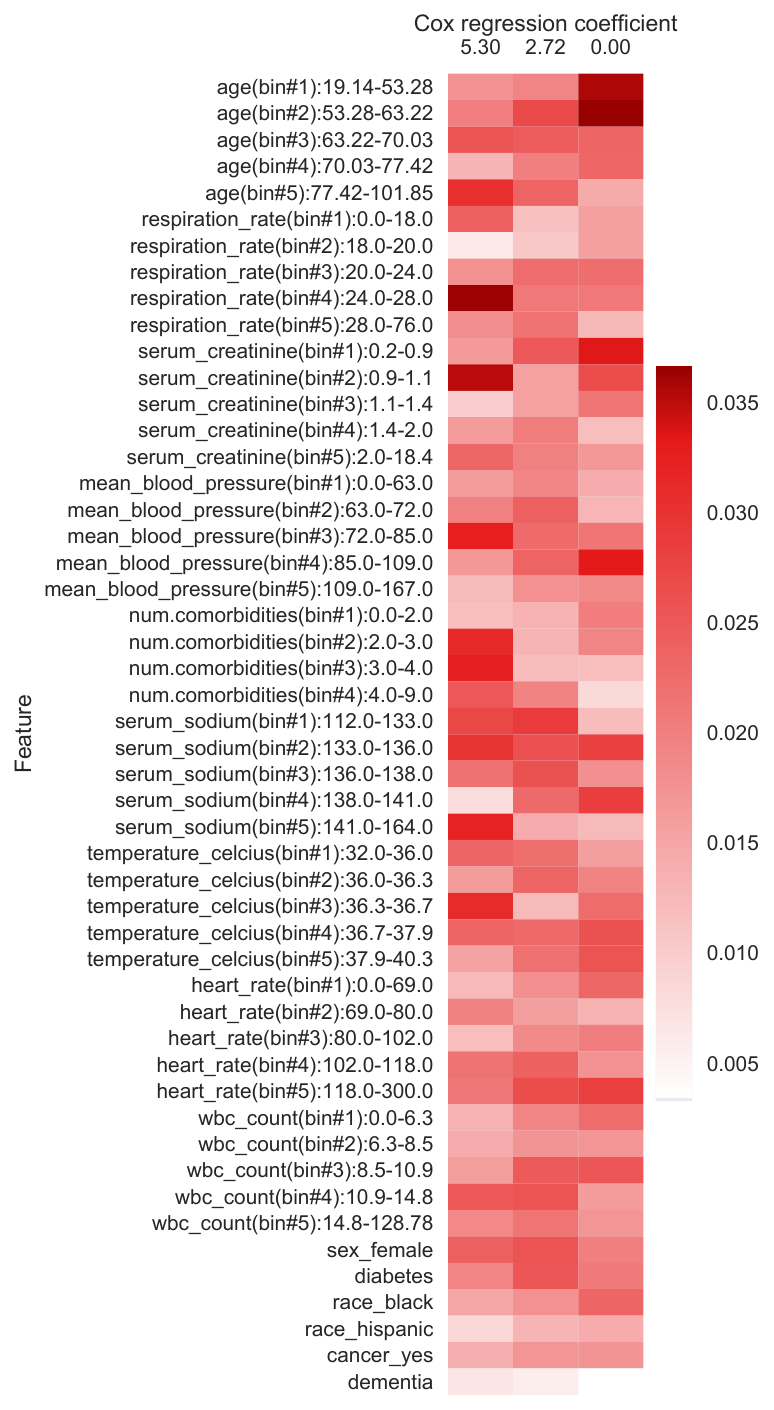}
\caption{Topics learned by \textsc{scholar lda-cox} on the \textsc{support2} (COPD/congestive heart failure/cirrhosis) dataset. Columns index topics and rows index features/``words''. The values are probabilities of each feature conditioned on being in a topic.}
\label{fig:support2-heatmap}
\end{figure}

\begin{figure}
\centering
\includegraphics[scale=0.7]{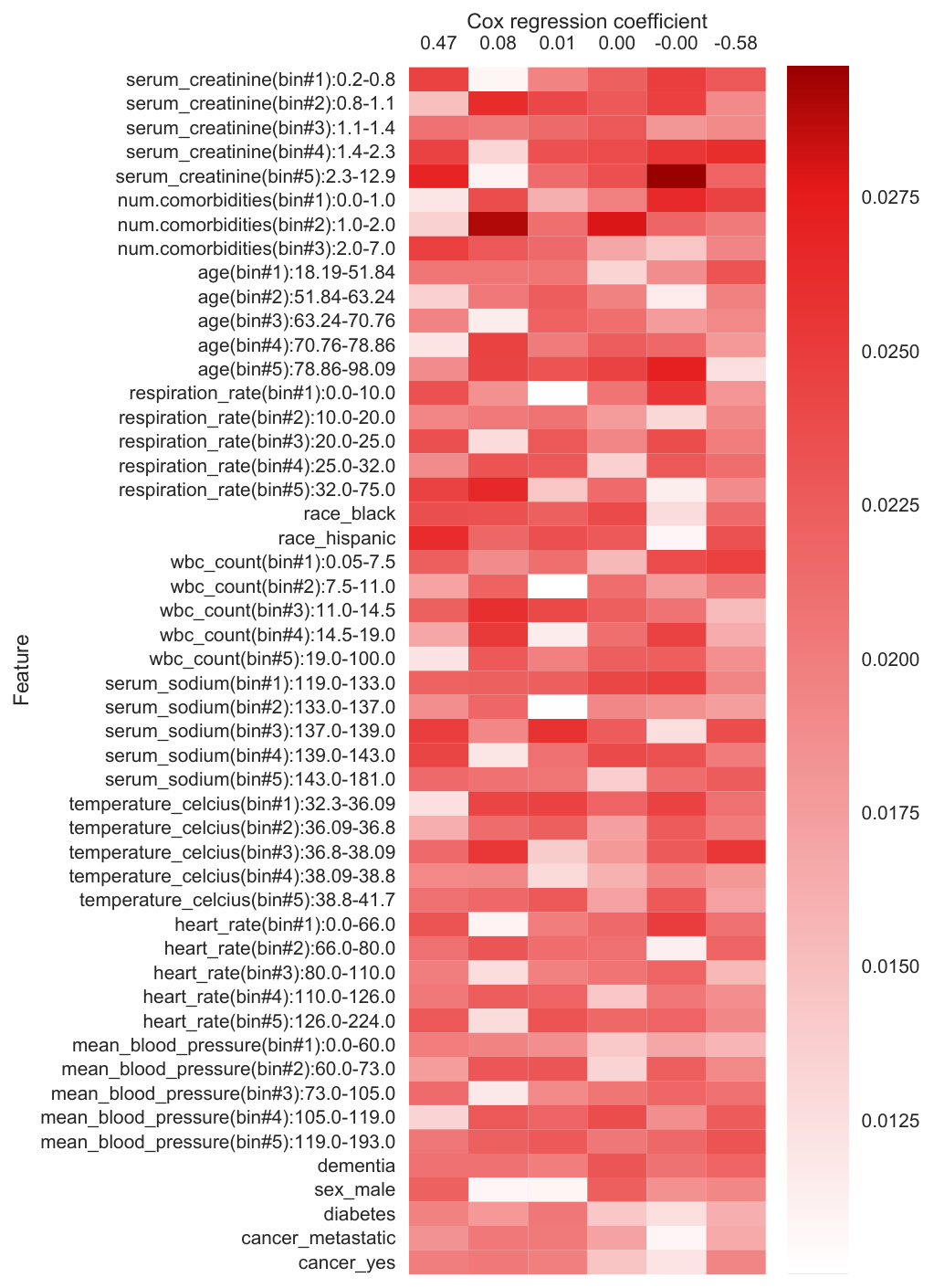}
\caption{Topics learned by \textsc{scholar lda-cox} on the \textsc{support4} (coma) dataset. Columns index topics and rows index features/``words''. The values are probabilities of each feature conditioned on being in a topic. Note that the Cox regression coefficient $-0.00$ actually corresponds to a value of $-0.00011$.}
\label{fig:support4-heatmap}
\end{figure}

\begin{figure}
\centering
\includegraphics[scale=0.45]{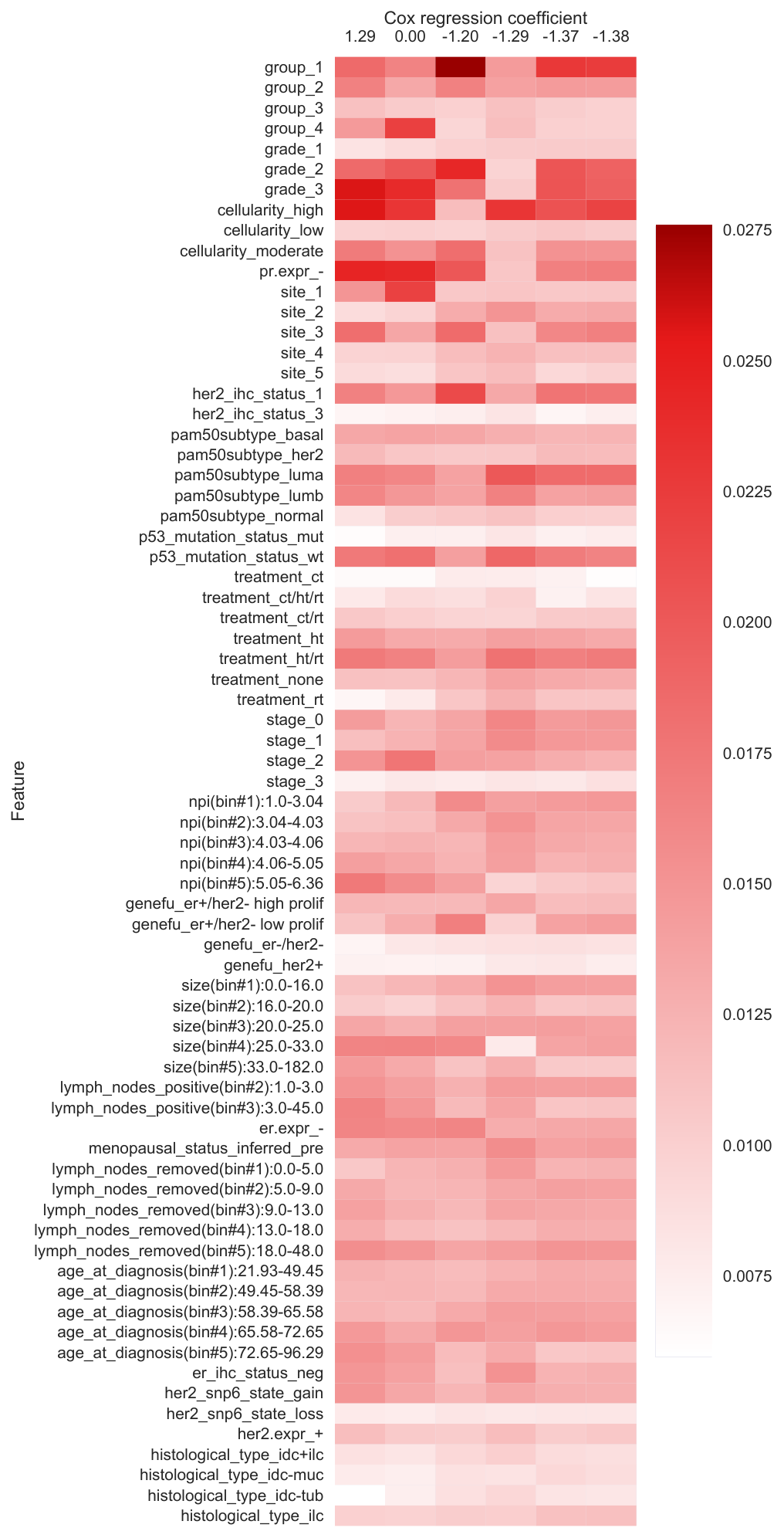}
\caption{Topics learned by \textsc{scholar lda-cox} on the \textsc{metabric} (breast cancer) dataset. Columns index topics and rows index features/``words''. The values are probabilities of each feature conditioned on being in a topic.}
\label{fig:metabric-heatmap}
\end{figure}

\begin{figure}
\centering
\includegraphics[scale=0.45]{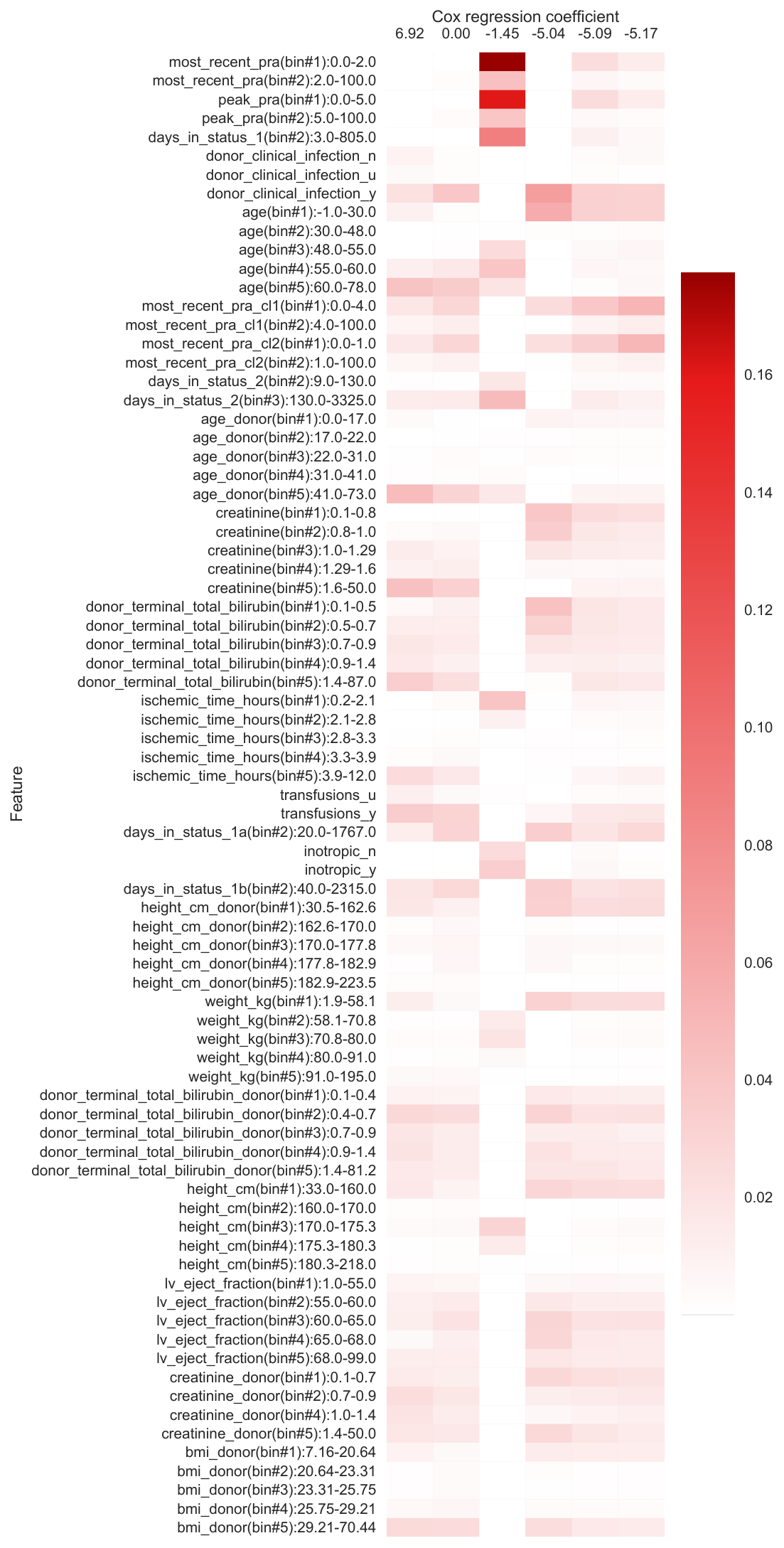}
\caption{Topics learned by \textsc{scholar lda-cox} on the \textsc{unos} (heart transplant) dataset. Columns index topics and rows index features/``words''. The values are probabilities of each feature conditioned on being in a topic.}
\label{fig:unos-heatmap}
\end{figure}

\begin{figure}
\centering
\includegraphics[scale=0.45]{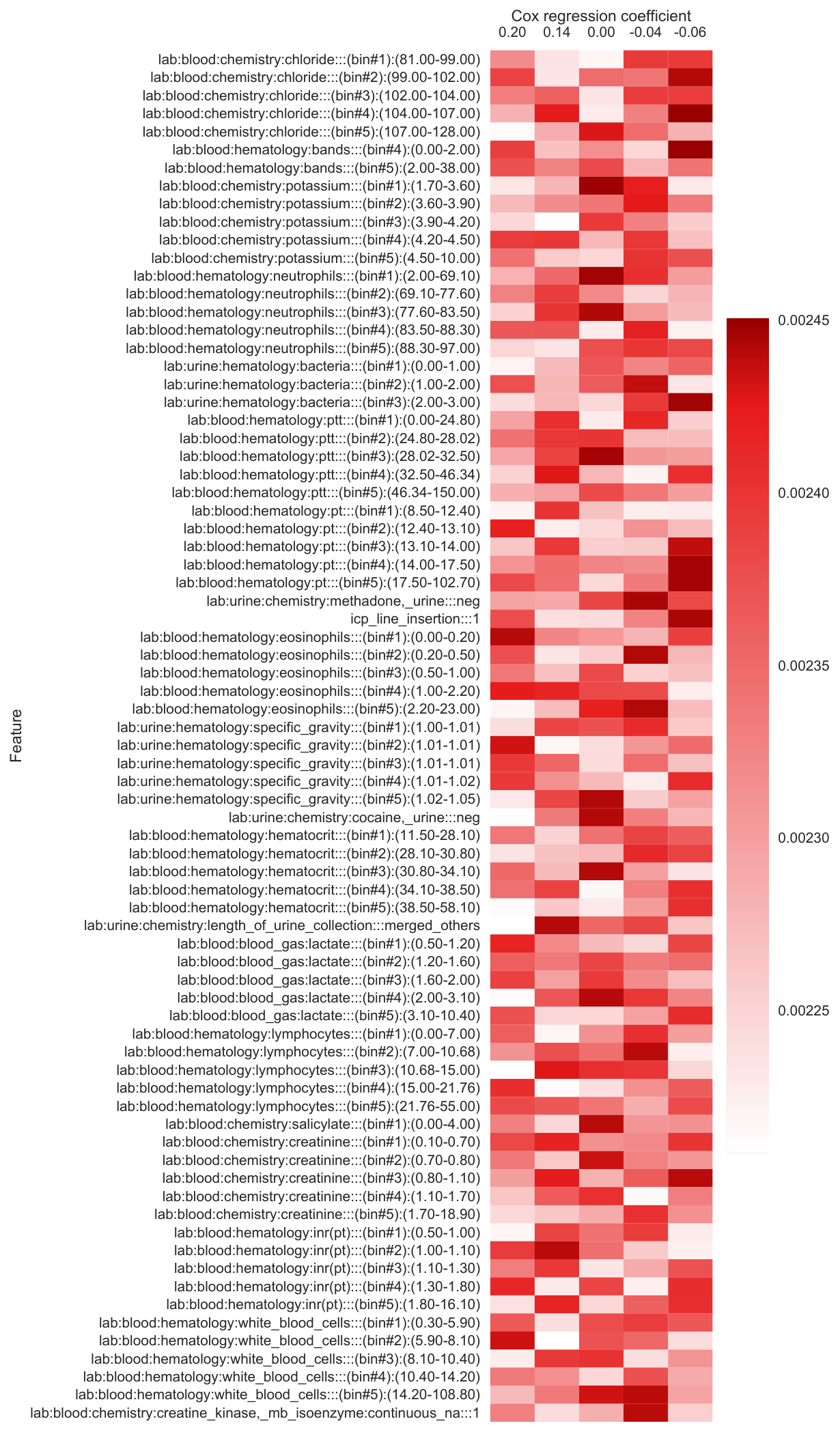}
\caption{Topics learned by \textsc{scholar lda-cox} on the \textsc{mimic-ich} (intracerebral hemorrhage) dataset. Columns index topics and rows index features/``words''. The values are probabilities of each feature conditioned on being in a topic.}
\label{fig:ich-heatmap}
\end{figure}

In producing these heatmaps, we also tried a few variations on the plots to present to a clinician. We sorted the words instead based on the largest difference between word probabilities across topics (i.e., rank words based on the score $(\max_{g=1,\dots,k} A_{g,v}) - (\min_{g=1,\dots,k} A_{g,v})$ for the $v$-th word) and also based on the average probability across topics ($\frac{1}{k}\sum_{g=1}^k A_{g,v}$). Qualitatively, we did not find an advantage to using these compared to the score we first presented of using the maximum word probability across topics. We also tried instead of using the raw word probabilities per topic, re-ranking words based on the topic TF-IDF score by \citet[equation (4.3)]{blei2009topic} and also based on the IDF score by \citet{alokaili2019re}. Qualitatively, we found that the topic TF-IDF weighting highlights a few words per topic but this weighting can be a bit too aggressive (the few words highlighted could be hard to interpret). IDF weighting could help draw out underrepresented words. Overall, though we did not see a clear advantage to using TF-IDF or IDF weighting in presenting the heatmap visualizations.

Note that prior to using our heatmap visualizations, we first tried providing a clinician with a listing of most probable words per topic. This is a standard approach for interpreting LDA models for text data. However, this way of conveying information turned out to be difficult for the clinician to quickly parse. For example, a feature might be in the top 20 most probable words for two different topics, and at that point understanding the difference in how probable the feature is across the two topics would be helpful. A listing of top words per topic did not make it easy to quickly find this information. For this reason, we switched to a heatmap visualization where each row of the heatmap directly gives us a quick way to compare probabilities of a feature/word across topics.

\endgroup

\section{Discussion}
\label{sec:discussion}

Despite many methodological advances in survival analysis with the help of deep learning, these advances have predominantly not focused on interpretability. Model interpretation can be especially challenging when there are many features and how they relate is unknown. In this paper, we show that neural survival-supervised topic models provide a promising avenue for learning structure over features in terms of ``topics'' that help predict time-to-event outcomes. These topics can be used by practitioners to check if learned topics agree with domain knowledge and, if not, to help with model debugging.

Our work thus far has a number of limitations. We discuss some of these limitations next.

\begin{figure}
\centering
\begin{subfigure}[b]{\linewidth}
\centering
\includegraphics[scale=.54]{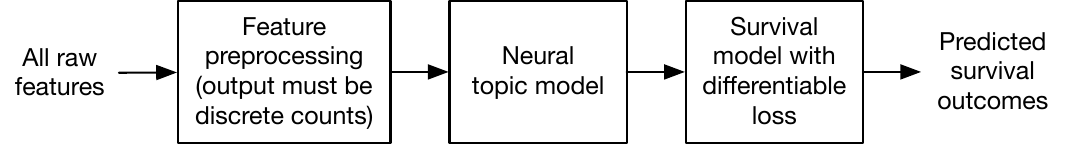}
\caption{}
\label{fig:framework}
\end{subfigure}

\begin{subfigure}[b]{\linewidth}
\centering
\includegraphics[scale=.54]{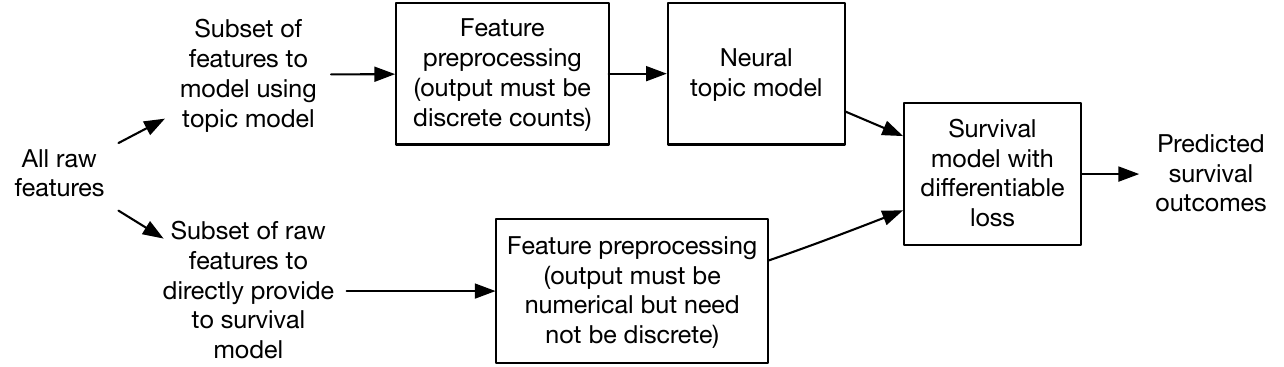}
\caption{}
\label{fig:intermediate-framework}
\end{subfigure}

\begin{subfigure}[b]{\linewidth}
\centering
\includegraphics[scale=.54]{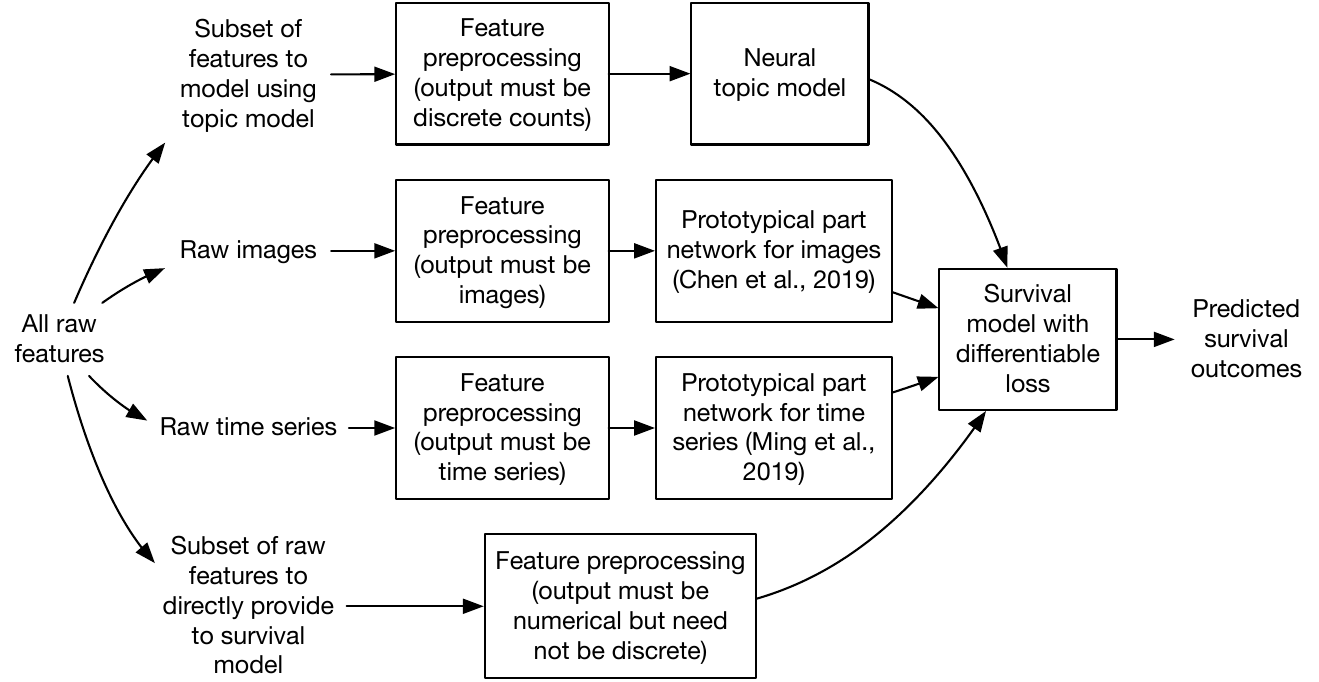}
\caption{}
\label{fig:generalized-framework}
\end{subfigure}

\caption{Incorporating different raw feature types: (a) our framework, (b) an extension of our framework allowing some raw features (which need not be discretized) to be directly modeled by the survival model, and (c) an extension of our framework that also uses prototypical part networks \citep{chen2019looks,ming2019interpretable} that are in some sense like topic models but for images and time series (we can omit different parts of this general framework depending on the raw input data that are available, e.g., if images are not available, then we remove the part involving prototypical part networks for images).}
\end{figure}

\paragraph{Moving beyond discrete data}
Our focus has been on when the raw features are encoded in a format specifying whether different historical clinically relevant events occur or not (the ``words'' of the topic model). This encoding inherently is discrete. The discretized raw counts then get modeled by a neural topic model, and the topics are treated as the input ``features'' for the survival model, as shown in Figure~\ref{fig:framework}. Discretizing continuous data inherently results in some loss in information. Better understanding how different discretization strategies (such as those described in Appendix~\ref{appendix:Other-Preprocessing}) impacts learned neural survival-supervised topic models in terms of accuracy and interpretability is an important direction for future research. Note that it is possible to also have some user-specified raw features be modeled directed by the survival model rather than being modeled by the topic model first, as shown in Figure~\ref{fig:intermediate-framework}; in this case, the raw features directly modeled by the survival model need not be discretized. For example, depending on the problem, we may want to have age be directly modeled by the survival model (e.g., a Cox model) rather than being explained by the topic model. As another example, consider gender being directly modeled by the survival model and not provided to the topic model. We could still try to understand how gender relates to the topics learned by adding interaction terms for the survival model (e.g., an indicator variable specifying whether female and topic 1 jointly occur, whether female and topic 2 jointly occur, etc).

Separately, much of the same ideas we presented in interpreting neural topic models readily apply to \emph{prototypical part networks} (ProtoPNets) \citep{chen2019looks,ming2019interpretable}, which behave like neural topic models but for raw data that are images or time series. Note that ProtoPNets can directly work with continuous-valued features without discretization. For example, given an input image, a ProtoPNet transforms the image into a vector representation specifying how much of each of $k$ different prototypes are present in the image (``similarity scores'' that are nonnegative); this vector representation behaves much like the topic weight vectors $W_i$'s that we have considered and could be fed as input to a survival model incorporating a background topic. Using these ideas, it is possible to build survival-supervised neural topic models that accept heterogeneous inputs, for example using the discrete ``words'' that we have considered in this paper, alongside images and time series (that could be left as continuous-valued). Of course, we could again choose some features to be directly modeled by the survival model. The overall diagram depicting this setup is shown in Figure~\ref{fig:generalized-framework}.

\paragraph{Incorporating additional structure in topics}
Topics learned by neural survival-supervised topic models vary in how easy they are for a clinician to interpret. We suspect that to improve interpretability, additional regularization is essential. For example, one possible research direction is to automatically find clinical measurements that do not plausibly co-occur within individual subjects, and add regularization that disallows these ``contradictory'' clinical measurements from both being highly probable within the same topic. For example, hematocrit and hemoglobin should be highly correlated, so we would expect that if a topic says one has a high probability of taking on a high value, then the topic should also say that the other has high probability of taking on a high value.

As another example, when a continuous measurement is discretized, we currently do not impose any constraints on the resulting discretized variables even though they are, of course, highly dependent on each other (i.e., a continuous variable is converted into a collection of variables that correspond to whether different discretization bins occur, and when one of them occurs, we know that the others cannot occur). A fix to this issue would be to add in loss terms to say when specific ``words'' explicitly do \emph{not} occur.

A less straightforward relationship to encourage is that a specific continuous variable (that has been discretized) should have a monotonic association with the survival time. Neither the raw continuous variable nor its corresponding discretized variables are provided directly as input to the survival model---instead they are treated as inputs to the topic model. One possible workaround is as follows. Suppose that we think age should have a monotonic association with survival time, and that it is discretized into bins 1 through 5, going from smaller to larger ages. Then for a specific topic, we could constrain the topic's probabilities for the discretized variables for age to be monotonic (i.e., the probabilities of the bins either increase from bin 1 up to bin 5, or they decrease from bin 1 up to bin 5 depending on whether we want the presence of the topic to be associated with higher or lower ages).

\paragraph{Topic stability}
As a separate direction that requires further investigation, thus far, we have not conducted experiments to quantify how ``stable'' the topics learned are across, for example, different random neural net parameter initializations. This is a problem more broadly found in training neural networks and is referred to as ``prediction churn'' \citep{bahri2021locally}. Better understanding how much the learned topics change due to random initialization would be helpful. We suspect that introducing regularization---such as the one we suggested for encouraging plausible co-occurrences---would lead to more stable topics learned. Even if we develop an improved understanding of topic stability, we would further need to understand how best to communicate this information to clinicians.

\paragraph{Competing risks}
In this paper, we focused on the standard right-censored survival analysis setup. We point out that our framework trivially extends to the competing risks setting, where we further want to reason about the cause of death (or more generally, a collection of competing critical events that could occur, where whichever occurs first prevents the other critical events from occurring). In this case, for each training subject, we assume that in addition to the subject's raw clinical events data, observed time, and indicator variable for whether death occurred, if death did occur, we also know the cause of death (among a finite set of causes under consideration). Standard competing risk models (e.g., see Chapter~8 of the textbook by \citet{kalbfleisch2002statistical}) can be used in place of the survival model in our neural net framework to obtain a neural topic model for competing risks. For example, one approach would be to have a Cox loss per cause of death, where the key idea here is that standard competing risk models still can be framed as minimizing a differentiable loss function (specifically a negative log likelihood). Empirically studying the resulting neural topic models for competing risks could provide interesting practical insights, with the goal of automatically surfacing feature relationships through a topic model, and finding associations between topics and the different causes of death.

\paragraph{Theoretical analysis}
Lastly, we mention that developing theory to understand when and why neural survival-supervised topic models work would be valuable. In particular, for what datasets should we expect to be able to learn such neural topic models that have sufficiently high prediction accuracy and are also easy to interpret? What special structure should be present in the data and how much data do we need? How does data preprocessing (e.g., discretization) impact these neural topic models? Finding theory that answers these questions could help clinicians understand when our proposed framework is most effective and what the best practices are in collecting and preprocessing data for use with our framework.

\subsubsection*{Acknowledgments} This work was supported in part by National Science Foundation CAREER award \#2047981, and by Health Resources and Services Administration contract 234-2005-370011C. The content is the responsibility of the author alone and does not necessarily reflect the views or policies of the National Science Foundation or the Department of Health and Human Services, nor does mention of trade names, commercial products, or organizations imply endorsement by the U.S.~Government.

\appendix

\section{Datasets and Preprocessing Details}
\label{appendix:data-preprocessing}

We describe the seven datasets we use and how we preprocess the data to obtain feature vectors of the format specified in Section~\ref{sec:background}.

\subsection{Datasets}

\paragraph{\textsc{support}}
The dataset from the Study to Understand Prognoses Preferences Outcomes and Risks of Treatment (SUPPORT) \citep{knaus1995support} is freely available online.\footnote{\url{http://biostat.mc.vanderbilt.edu/wiki/Main/SupportDesc}} This dataset contains clinical features collected from seriously ill hospitalized adults, such as their age, presence of cancer, and neurologic function. These features were collected from patients on the third day after the study started, and patients were followed for survival up to 5.56 years after entering the study. We do not use all the features and instead use the same 14 features that were used by \citet{katzman2018deepsurv} in their experiments. We further split the dataset into four datasets corresponding to different disease groups (acute respiratory failure/multiple organ system failure, cancer, coma), as done by \citet{harrell2015regression}. After we created these four subsets, all subjects from the cancer group have identical values for a clinical feature related to cancer presence, so this feature was removed only for the cancer cohort, resulting in 13 clinical features for the \textsc{support3} dataset. Furthermore, of the 14 features used, two features (creatinine and white blood count) had suggested imputation values from the official SUPPORT documentation that we used. After using this imputation, any data point that still had missing values for any of the 14 features used was omitted, resulting in the dataset sizes given in Table~\ref{tab:datasets}.

\paragraph{\textsc{metabric}}
The Molecular Taxonomy of Breast Cancer International Consortium (METABRIC) dataset is available on the Synapse platform\footnote{\url{https://www.synapse.org/}}. This dataset contains clinical and genetic features from breast cancer patients, and their respective survival durations. We only used a subset of 24 features that are available for open use through Synapse. This dataset includes 1981 breast cancer patients in total, around 55.2\% of whom were censored and not followed until death. The original METABRIC paper \citep{curtis2012genomic} discusses how the dataset's clinical features were defined in more detail. 

\paragraph{\textsc{unos}}
The UNOS dataset was extracted from the United Network for Organ Sharing (UNOS) database\footnote{\url{https://www.unos.org/data/}}, and curated in order to replicate the pre-processing documented by \citet{lee2018deephit} and \citet{yoon2018personalized}. We selected only patients who went through heart transplantations in the 30-year window from January 1985 to December 2015. Because \citet{yoon2018personalized} did not document the exact list of feature names that we could directly extract from the database, we attempted to the best of our ability to curate a list of features that overlaps the most with the feature table presented by them. We ended up with 49 features in total, among which 31 are recipient-related, 12 are donor-related, 6 are compatibility related. For this dataset, our objective is to predict patients' post-transplantation survival time. Because we assumed December 2015 to be the end of data collection, patients who were still alive as of December 2015 are all considered censored samples. Among 62644 patients who underwent transplantation, around 50.2\% are censored samples.

\paragraph{\textsc{mimic-ich}}
The intracerebral hemorrhage (ICH) dataset we evaluated on is created from MIMIC-III (version 1.4), a critical care health records database containing 52 thousand individuals and their hospital encounters involving admission to the ICU at Beth Israel Deaconess center between 2001 and 2012 \citep{johnson2016physionet,johnson2016mimic}. Experiments were conducted using a subset of the MIMIC-III data consisting of patients having spontaneous intracerebral hemorrhage requiring admission to the ICU. Patients were included in the study if they have an ICU admission with a primary billing code of intracerebral hemorrhage, resulting in a cohort of 961 individuals. For patients who are admitted to the ICU multiple times, we only consider their first visit to the ICU within the dataset. We aim to predict patients' lengths of stay in the ICU (specifically time until discharge). This subset of the data has no right-censoring in the sense of data no longer being collected midway through a patient's ICU stay. However, 23.1\% of the patients die in the ICU; for these individuals, we record the time until death as the observed time and set the indicator variable for whether the patient is discharged to 0. In particular, death is effectively treated as the sole censoring event.

Features extracted include demographics, medications, billing codes, procedures, laboratory measurements, events recorded into charts, and vitals. Features were extracted from the relational database into a 4-column format for \emph{patient id}, \emph{time}, \emph{event}, and \emph{event value}. To prevent erroneous merging of different events into a single event, and to provide more informative events, event strings are concatenations of the event descriptor prefixed with the table from which they are derived and additional relevant information such as measurement type, measurement units, etc. Because events recorded in charts are sometimes automated and sometimes manually entered, a physician-developed mapping and lower-casing all fields were used to resolve duplicate entries. As we aim to predict the patient length of stay in ICU, we extract clinical events from the subjects' electronic health records strictly before ICU admission. After preprocessing, the total number of features used for prediction is 1530.

\subsection{Features Used}
\label{appendix:Features-Used}

For all of our datasets, categorical features were one-hot encoded. Specifically to the Cox proportional hazards and lasso-regularized Cox baselines, for each categorical feature, one category was removed as the reference column. For methods that use topic modeling, we realized it does not make sense to encode numeric clinical events as they are. Instead, numeric clinical events were treated as categorical by mapping observed values to equally spaced ranges by quintile (5 bins of roughly equal number of subjects per bin). When values of a numeric clinical event are highly cluttered (i.e., the 20/40/60/80 percentile values of the event do not correspond to 4 unique threshold values so that there end up being fewer than 5 bins), we allow the number of bins to be less than 5, where the resulting bins can have imbalanced numbers of subjects. For instance, if there are fewer than 5 unique values for the clinical event across data points, then we cannot discretize the event into 5 nonempty bins.

Features for the \textsc{mimic-ich} dataset were created slightly differently.
Our definition of clinical events mean that a subject can have multiple instances of one event; for example, one patient might have multiple results for a particular lab test on file. Under this case, single-occurrence categorical events (e.g., gender) were one-hot encoded as usual; multiple-occurrence categorical events (e.g., urine color) were encoded by counting each category's occurrences in a single subject's records. For numeric clinical events, as a subject may have a list of numeric values recorded, we engineered numeric features that captured the minimum, maximum, median, and length of a subject's list of recordings. However, this was not necessary for methods that use topic modeling, because mapping values to equally spaced bins took care of multiple-occurrence numeric events for us.

We would also like to note that missing records were not imputed as missing certain events can have clinical significance. Therefore, for features with incomplete records, the missing entries were first filled with zeros, and then an additional feature was added solely to indicate whether missingness is observed for each subject; this approach to handling missing data is motivated by the work of \citet{lipton2016modeling}. While we added features that solely indicate missingness for all baseline methods, methods that use topic modeling do not require encoding missingness explicitly. For topic modeling based methods, feature vectors encode number of occurrences, so a patient with missing feature simply has that feature's number of occurrences set to~0. For this reason, we did not explicitly encode missingness as a separate feature for methods that use topic modeling.

\subsection{Other Possible Ways to Encode Clinical Measurements}
\label{appendix:Other-Preprocessing}

Our feature preprocessing has largely been chosen to be relatively easy to explain. We now mention other strategies that are possible for discretization and, separately, for summarizing a feature across time.

\paragraph{Discretization} We discretize continuous features into quintiles (as we mentioned earlier, sometimes this is not possible so we simply use fewer than 5 bins), which is a simple strategy that can be used for different continuous features without a priori knowledge. However, if one did have domain knowledge about how specific features could be discretized, then such discretization strategies could be used instead of the simple quintile binning strategy. As an example, there are specific cutoffs whereby cohorts are defined (e.g., lactate levels of 4), and where medical interventions are indicated (e.g., mean arterial pressures below 65).

Alternatively, one could even learn how to discretize a specific continuous feature (a single real number). For instance, taking the feature's value across the training data, we could use a user-specified clustering algorithm (e.g., Jenks natural breaks \citep{jenks1967data}) to cluster on the observed values of the continuous feature to decide on how to discretize (the thresholds could come from the boundary points between clusters). A different strategy is to learn a decision tree for survival analysis using a single continuous feature across the data. Such a tree could be learned greedily (using the same tree learning strategy as in random survival forests \citep{ishwaran2008random}) or optimally by solving a mixed-integer program \citep{bertsimas2022optimal}: the leaves of the learned tree directly correspond to the discretization bins. A generalization of this idea is possible in which multiple continuous features could be discretized together (train a single decision tree with these different features and then let the final tree leaves correspond to the discretization bins).

\paragraph{Summarizing a feature across time} For ease of exposition, we had simply counted how often a feature occurred across time to obtain the raw counts matrix $X$. If we had domain knowledge of how a specific feature should be summarized across time, then we could take this into account when summarizing the feature. For example, if we take many oxygen saturation measurements within a few minutes, clinically it is common to take the highest measured value because the physiologic process prevents rapid fluctuations in saturation, and the measurement is intended to grossly assess oxygenation and perfusion. Alternatively, we could use the approach by \mbox{\citet{johnson2021learning}} that automatically learns how to summarize continuous or discrete features across time in such a way that the summary features are clinically interpretable. Each summary feature can then be discretized using any user-specified discretization strategy, such as the clustering or decision-tree approaches we described in the previous paragraph.

\section{Hyperparameter Search}
\label{appendix:hyperparameter-grids}

We use grid search, with the same grid of hyperparameters used across datasets per model as given in Table~\ref{tab:hyperparameter-grids}. For neural net approaches, we always train using Adam \citep{kingma2015adam} with a batch size of 256 and use early stopping (no improvement in best validation $C^{\textsf{td}}$ index within 10 epochs) with a budget of 512 epochs; however we do vary the learning rate and sweep over the choices of 0.01 and 0.001.

\begin{table}
\centering
\caption{Hyperparameter grids used during model training.}
\label{tab:hyperparameter-grids}
\footnotesize
{\renewcommand{\arraystretch}{1.2}
\begin{tabular}{cc}\toprule
Model & Hyperparameter Grid \\\midrule
\textsc{cox} & \parbox{0.8\linewidth}{lasso regularization weight: 0, 0.0001, 0.001, 0.01, 0.1, 1.0} \\
\cmidrule[0.01em](l{.75em}r{.75em}){1-2}
\textsc{rsf} & \parbox{0.8\linewidth}{number of trees: 100 \\
number of features used per split: sqrt of total number of features, rounded up \\
max depth: 2, 4, 6, 8} \\
\cmidrule[0.01em](l{.75em}r{.75em}){1-2}
\textsc{deepsurv} & \parbox{0.8\linewidth}{number of hidden layers for the multilayer perceptron: 1, 2, 4 \\
number of nodes per hidden layer: 16, 32, 64} \\
\cmidrule[0.01em](l{.75em}r{.75em}){1-2}
\textsc{deephit} & \parbox{0.8\linewidth}{number of hidden layers for the multilayer perceptron: 1, 2, 4 \\
number of nodes per hidden layer: 16, 32, 64 \\
number of durations (in time discretization): 64, 128 \\
$\alpha$ (in original DeepHit paper; not LDA Dirichlet hyperparameter): 0.1, 0.5, 0.9 \\
$\sigma$ (in original DeepHit paper; not AFT scale parameter): 0.1, 1.0, 10.0} \\
\cmidrule[0.01em](l{.75em}r{.75em}){1-2}
\textsc{naive lda-cox} & \parbox{0.8\linewidth}{number of topics: 2, 3, 4, 5, 6} \\
\cmidrule[0.01em](l{.75em}r{.75em}){1-2}
\textsc{scholar lda-cox} & \parbox{0.8\linewidth}{number of topics: 2, 3, 4, 5, 6 \\
word embedding dimension: 16, 32, 64 \\
$\lambda_{\textsf{survival}}$: 1, 100, 10000, 1000000} \\
\cmidrule[0.01em](l{.75em}r{.75em}){1-2}
\textsc{scholar lda-aft} & \parbox{0.8\linewidth}{number of topics: 2, 3, 4, 5, 6 \\
word embedding dimension: 16, 32, 64 \\
$\lambda_{\textsf{survival}}$: 1, 100, 10000, 1000000 \\
$\lambda_{\textsf{ranking}}$: 1} \\
\cmidrule[0.01em](l{.75em}r{.75em}){1-2}
\textsc{scholar sage-cox} & \parbox{0.8\linewidth}{number of topics: 2, 3, 4, 5, 6 \\
word embedding dimension: 16, 32, 64 \\
$\lambda_{\textsf{survival}}$: 1, 100, 10000, 1000000 \\
$\lambda_{\textsf{small-deviation}}$: 0.005, 0.05, 0.5, 5} \\
\cmidrule[0.01em](l{.75em}r{.75em}){1-2}
\textsc{scholar sage-aft} & \parbox{0.8\linewidth}{number of topics: 2, 3, 4, 5, 6 \\
word embedding dimension: 16, 32, 64 \\
$\lambda_{\textsf{survival}}$: 1, 100, 10000, 1000000 \\
$\lambda_{\textsf{ranking}}$: 1 \\
$\lambda_{\textsf{small-deviation}}$: 0.005, 0.05, 0.5, 5} \\
\bottomrule
\end{tabular}
}
\end{table}

\bibliographystyle{plainnat}
\bibliography{survival_topic_models}

\end{document}